\pdfoutput=1

\documentclass[11pt]{article}

\usepackage[dvipsnames]{xcolor}
\usepackage{tcolorbox}\tcbuselibrary{skins}
\usepackage{makecell}

\definecolor{CB_lightCyan}{HTML}{99DDFF}
\definecolor{CB_pear}{HTML}{BBCC33}
\definecolor{CB_pink}{HTML}{FFAABB}

\usepackage{todonotes}

\usepackage[obeyspaces]{url}
\usepackage{hyperref}
\newcommand{\prettyUrl}[2]{\href{#1}{\path{#2}}}

\newcommand{\fon}[1]{\fontfamily{#1}\selectfont} 

\tcbset{entailmentPrompt/.style={
    enhanced,
    size=fbox,
    boxrule=2pt,
    arc=2mm,
    auto outer arc,
    left=1pt,
    right=1pt,
    top=1pt,
    bottom=1pt,
    fontupper=\fon{cmtt}, 
    colback=CB_lightCyan!15,
    colframe=CB_lightCyan!30,
    coltitle=CB_lightCyan!25!black, 
}}

\tcbset{taggingPrompt/.style={
    enhanced,
    size=fbox,
    boxrule=2pt,
    arc=2mm,
    auto outer arc,
    left=1pt,
    right=1pt,
    top=1pt,
    bottom=1pt,
    fontupper=\fon{cmtt}, 
    colback=CB_pear!15,
    colframe=CB_pear!30,
    coltitle=CB_pear!25!black, 
}}

\tcbset{otherText/.style={
    enhanced,
    size=fbox,
    boxrule=2pt,
    arc=2mm,
    auto outer arc,
    left=1pt,
    right=1pt,
    top=1pt,
    bottom=1pt,
    fontupper=\fon{cmtt}, 
    colback=CB_pink!15,
    colframe=CB_pink!30,
    coltitle=CB_pink!25!black, 
}}
\usepackage[preprint]{acl}

\usepackage{times}
\usepackage{latexsym}

\usepackage[T1]{fontenc}

\usepackage[utf8]{inputenc}

\usepackage{microtype}

\usepackage{inconsolata}

\usepackage{graphicx}

\usepackage{array}
\usepackage{multirow}
\usepackage{graphicx}
\usepackage{booktabs}
\usepackage{pifont}
\usepackage{xspace}

\definecolor{dcolor}{RGB}{193, 167, 214}
\definecolor{ccolor}{RGB}{166, 209, 233}
\definecolor{scolor}{RGB}{227, 227, 227}
\definecolor{pcolor}{RGB}{255, 227, 183}
\definecolor{hotpink}{RGB}{255, 105, 180}

\usepackage{soul}

\definecolor{vanilla}{rgb}{0.95, 0.9, 0.67}
\DeclareRobustCommand{\hlcolor}[1]{{\sethlcolor{vanilla}\hl{#1}}}
\DeclareRobustCommand{\phlcolor}[1]{{\sethlcolor{pcolor}\hl{#1}}}
\DeclareRobustCommand{\dhlcolor}[1]{{\sethlcolor{dcolor}\hl{#1}}}
\DeclareRobustCommand{\chlcolor}[1]{{\sethlcolor{ccolor}\hl{#1}}}
\DeclareRobustCommand{\shlcolor}[1]{{\sethlcolor{scolor}\hl{#1}}}

\newcommand{\apple}{\raisebox{-2pt}{\includegraphics[height=1em]{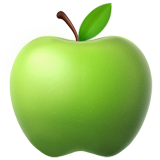}}}
\newcommand{\fish}{\raisebox{-2pt}{\includegraphics[height=1em]{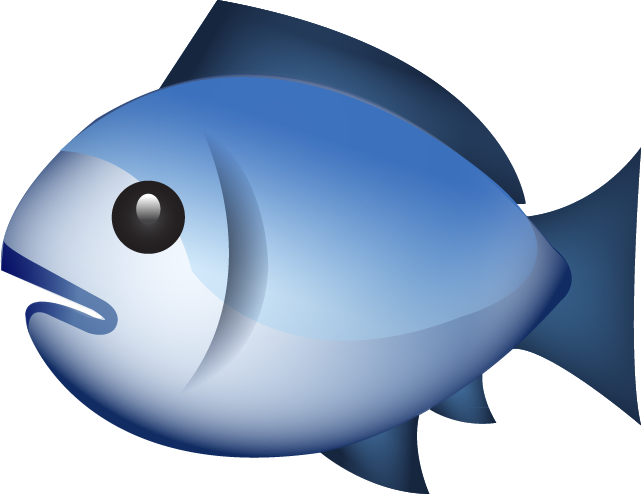}}}

\newcommand{\github}{\raisebox{-1.5pt}{\includegraphics[height=1.05em]{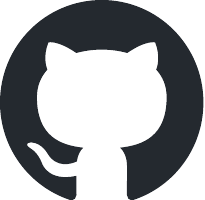}}\xspace}
\newcommand{\huggingface}{\raisebox{-1.5pt}{\includegraphics[height=1.05em]{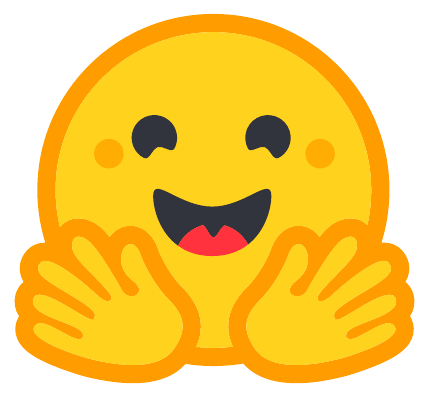}}\xspace}
\newcommand{\atc}{\texttt{ATC}~\apple}
\newcommand{\mathfish}{\texttt{MathFish}~\fish}

\title{\mathfish: Evaluating Language Model Math Reasoning via\\Grounding in Educational Curricula}

\author{Li Lucy\textsuperscript{1,2} \quad Tal August\textsuperscript{1} \quad Rose E. Wang\textsuperscript{3}  \\
\textbf{Luca Soldaini\textsuperscript{1}} \quad \textbf{Courtney Allison\textsuperscript{4}} \quad \textbf{Kyle Lo\textsuperscript{1}}
\vspace{0.3em} \\
  \textsuperscript{1}Allen Institute for AI \quad
  \textsuperscript{2}University of California, Berkeley\\
  \textsuperscript{3}Stanford University \quad
  \textsuperscript{4}EdReports
  \vspace{0.3em} \\
  \texttt{lucy3\_li@berkeley.edu} \quad \texttt{kylel@allenai.org}
}

\begin{document}
\maketitle
\begin{abstract}
To ensure that math curriculum is grade-appropriate and aligns with critical skills or concepts in accordance with educational standards, pedagogical experts can spend months carefully reviewing published math problems.
Drawing inspiration from this process, our work presents a novel angle for evaluating language models' (LMs) mathematical abilities, by investigating whether they can discern skills and concepts enabled by math content. 
We contribute two datasets: one consisting of 385 fine-grained descriptions of K-12 math skills and concepts, or \textit{standards}, from Achieve the Core (\atc), and another of 9.9K math problems labeled with these standards (\mathfish). 
We develop two tasks for evaluating LMs' abilities to assess math problems: (1) \emph{verifying} whether a problem aligns with a given standard, and (2) \emph{tagging} a problem with all aligned standards.
Working with experienced teachers, we find that LMs struggle to tag and verify standards linked to problems, and instead predict labels that are close to ground truth, but differ in subtle ways. 
We also show that LMs often \emph{generate} problems that do not fully align with standards described in prompts, suggesting the need for careful scrutiny on use cases involving LMs for generating curricular materials.
Finally, we categorize problems in GSM8k using math standards, allowing us to better understand why some problems are more difficult to solve for models than others.
\end{abstract}

\begin{center}
\small
\renewcommand{\arraystretch}{1.2}
\begin{tabular}{p{.03\columnwidth}p{.1\columnwidth}p{.65\columnwidth}}
    \github & \textbf{Code} & \href{https://github.com/allenai/mathfish}{\path{github.com/allenai/mathfish}} \\
    \huggingface & \textbf{Dataset} & 
    \href{https://huggingface.co/datasets/allenai/mathfish}{\path{hf.co/datasets/allenai/mathfish}}\\
\end{tabular}
\end{center}

\section{Introduction}

When assessing mathematical reasoning in large language models (LMs), a common approach is to test their problem solving abilities. Math is a popular domain for model evaluation \citep{cobbe2021training, hendrycks2021measuring, zhang2024careful}, as problem instances can be designed to target specific abilities. However, many datasets contain only coarse-grained information on what skills each problem assesses, such as general arithmetic or operation types \cite{hase2024unreasonable}. 
In practice, curricular experts’ categorizations of math are fine-grained, mapping materials to specific \textit{skills}, such as multiplication procedures for fractions, or \textit{concepts}, such as area and volume. Our work bridges this gap and examines a novel angle for evaluating LMs' mathematical understanding. We ask, how well can models identify specific skills and concepts that students learn or practice when completing math problems?

First, we contribute English datasets of 9.9K human-written math problems (\mathfish) scaffolded by 385 K-12 U.S. curriculum standards in Achieve the Core (\atc). These standards are informed by human learning progressions, and commonly used in real-world reviews of math content. In education, materials have focused \textit{alignment} with a standard if they enable students to learn the \textit{full intent} of concepts/skills described by that standard.\footnote{\url{https://achievethecore.org/page/2730/aligned-instructional-practice}} Identifying alignment can thus inform educators whether a set of materials adequately targets core learning goals for students.

Second, we provide a fine-grained assessment of LMs' abilities to reason about math pedagogy, skills, and concepts, by asking them to recognize whether a standard aligns with a given problem (\S\ref{sec:entail}, \S\ref{sec:tag}).
We experiment with two task formats: one in which we \textit{verify} whether a single standard aligns with a problem, and another in which we \textit{tag} each problem with any standards. 
Our experiments demonstrate that models achieve reasonable accuracy, but are not yet at expert-level performance across both task formats. 

Third, to further illustrate the utility of these task formats, we apply the best-performing models and prompting approaches on two case studies. In one, we work with K-12 math teachers with curriculum review experience to apply verification on LM-generated problems (\S\ref{sec:v2}). We find that a GPT-4 verifier tends to overestimate full standards alignment of generated problems compared to teachers. In the other, we tag GSM8k \cite{cobbe2021training}, a widely used grade school math dataset, with standards (\S\ref{sec:v3}). We find that GSM8k only covers around a third of all K-12 standards, and that LMs struggle more to solve problems tagged with higher grade levels’ standards. 

As LMs are deployed in more real-world use cases, it is increasingly important to evaluate models with potential users. Throughout this paper, we work with curriculum specialists and teachers to center their voices when evaluating LM capabilities. 
These educational experts informed us that identifying math standards in curricular materials is time-intensive, and reviewing a set of published materials can take 6-8 months to complete. 
This motivates us to investigate whether models can support reviewers by combing through subtle differences across standards. 
Altogether, we hope that these datasets, along with the tooling we contribute for transforming data into task formats for LMs, can facilitate a more granular understanding of LM reasoning around skills and concepts in math content.  

\begin{figure*}[t]
\centering
\includegraphics[width=\textwidth]{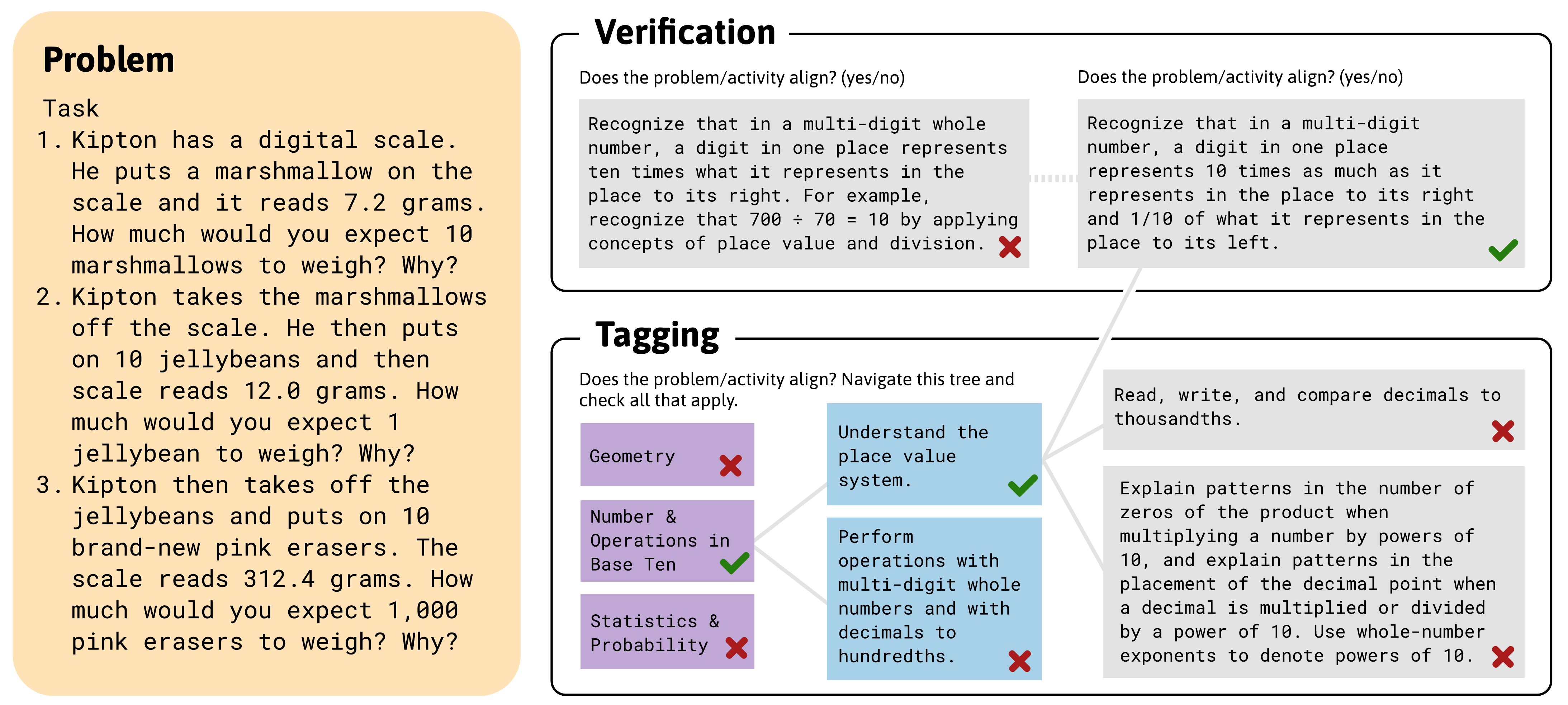}  
\caption{An example of a \mathfish{} \phlcolor{problem}, along with \dhlcolor{domains ($\mathcal{D}$)}, \chlcolor{clusters ($\mathcal{C}$)}, and \shlcolor{standards ($\mathcal{S}$)} it does and does not align with. Solid lines indicate hierarchical relationships, while a dashed line links conceptually connected standards. In addition, this figure illustrates two task formats: verification (\S\ref{sec:entail}) and tagging (\S\ref{sec:tag}). 
}
\label{fig:v1}
\end{figure*}

\section{Related Work}

\paragraph{Math benchmarks for evaluating LMs.}
Past work assessing mathematical reasoning capabilities of LMs have mainly focused on grade school arithmetic and algebra and problems that are easy to validate. These datasets may be written by annotators without formal pedagogical training \citep[e.g.][]{cobbe2021training, patel2021nlp, amini2019mathqa, ling2017program, miao2021diverse},
synthetically generated by LMs \citep{mitra2024orca, liu2023tinygsm}, and derived from advanced math competitions \citep{hendrycks2021measuring}. 
In contrast, our work characterizes real-world math curricula within an expansive, fine-grained taxonomy of math skills and concepts (\S\ref{sec:data}). 
Concurrent work by \citet{mishra2024mathcamps} presents an approach for generating problems that align with math standards, but targets a limited set of easily validatable standards. 
Our work instead focuses on the task of labeling problems across 300+ comprehensive and challenging K-12 standards, and emphasizes ecological validity by working with educators (\S\ref{sec:v2}). Earlier work in education has identified math standards in problems with models such as SentenceBERT \cite{li2024aligning}, task-adapted BERT \cite{shen2021classifying}, and support vector machines \cite{karlovvcec2012knowledge}. Our work re-envisions this task into a format suitable for evaluating instruction-tuned LMs. 

\paragraph{LMs for improving math education.}
LMs have increasing potential for improving math education through applications like automated feedback to educators \citep{jacobs2022promoting, demszky2023can, wang-demszky-2023-chatgpt},
automated tutoring \citep{hobert2019say, wang-etal-2024-bridging},
and lesson planning \citep{10.1145/3657604.3664698, kasneci2023chatgpt}.
Evaluating models with education domain experts before real-world deployment is crucial to best align with educational needs and mitigate potential harms \citep{patikorn2019generalizability,li2024aligning,lee2024math}. 
In our setting, we work with domain experts in a manner that centers real-world educational needs (\S\ref{sec:v2}). 

\paragraph{Evaluating LMs amid fine-grained differences.}
Our emphasis on the granularity of models' decisions relates to other work that focuses on small yet impactful differences in language. Examples include the generation of good distractors that sufficiently challenge question solvers \cite{Gao2018GeneratingDF, zesch-melamud-2014-automatic}, the mining of ``hard negatives'' to supervise models to discern positives and close negatives \cite{robinson2021contrastive, kalantidis2020hard}, and minimal modifications of model inputs that change ground truth labels \cite{gardner-etal-2020-evaluating}. In our case, our ``hard negatives'' and distractors are not artificially constructed, but instead originate from inter-standard relationships drawn by pedagogical experts.

\section{Grounding Math Content in Expert-Designed Standards}\label{sec:data}

\subsection{Task Definition \& Formats}\label{sec:task_formats}

What are task formats that both (1) evaluate LM ability to discern math skills or concepts in problems, as well as (2) reflect real-world usage patterns of LMs by pedagogical experts?
In the initial phase of this work, we met with professional curriculum reviewers to better understand their processes and definitions for what does (or does not) constitute alignment to an educational standard.
We identified two usage patterns in which these experts would employ language model assistance for performing curricular alignment---\emph{verifying} whether a problem aligns to a given standard and \emph{tagging} a problem with all aligned standards.

When \emph{verifying}, 
we check a single problem against a single standard. 
This may arise when one is confirming publishers' claims of alignment (\S\ref{sec:verif_exp}), or when evaluating whether LM-generated materials follow the standard indicated in prompt instructions (\S\ref{sec:v2}). This task format is structured as a binary yes/no question (Figure~\ref{fig:v1}): does a problem fully align with a given standard? Taking inspiration from textual entailment~\citep{dagan-entailment} and claim verification~\citep{thorne-etal-2018-fever}, this format also allows us to investigate models' sensitivity to narrowing differences in positive and negative standards when we perturb them in queries. 

Though verification is useful for checking problem alignment with a \emph{single} standard, reviewers are also often tasked with identifying \emph{all} aligned standards for a given problem.
In \emph{tagging}, 
we take a single problem and select all aligned standards from a provided set of candidates.
Prompting language models with hundreds of candidate standards
risks hitting context limits or models getting ``lost in the middle'' \citep{Liu2023LostIT}.
Instead, we take advantage of a \textit{domains}-\textit{clusters}-\textit{standards} tree structure that experts use to organize standards into a hierarchy; related standards are grouped as leaves on this tree (Figure~\ref{fig:v1}).
We define the tagging task as follows: given a math problem, (1) start at the top of the tree and select the best domain/s $\mathcal{D}$ it teaches, (2) then traverse to each selected $\mathcal{D}$s' subtrees, (3) repeat selection for cluster/s $\mathcal{C}$, (4) stop traversal upon reaching standards $\mathcal{S}$ at the leaves, and (5) tag the problem with selected $\mathcal{S}$. 
By traversing this tree, our tagging task format allows us to see whether models can make increasingly granular distinctions among adjacent concepts/skills. 

Across these task formats, we experiment with three models: Mixtral-8x7B~\citep{Jiang2024MixtralOE}, Llama-2-70b-chat~\citep{touvron2023llama}, and GPT-4-Turbo~\citep{Achiam2023GPT4TR}. We unify them under a single wrapper, which handles input truncation and API calls (Appendix~\ref{sec:appdx_model_wrap}). 

\subsection{Math Standards \& Organization}\label{sec:standards}
To facilitate these experiments, we introduce two datasets: Achieve the Core (\atc) describes math standards and their organization, and \mathfish{} links standard labels to problems (Table~\ref{tbl:data_stats}). Both involve Common Core State Standards (CCSS), which offer fine-grained and comprehensive coverage of K-12 math skills/concepts \cite{porter2011cc}.

\atc{}
consists of CCSS standard labels (e.g. \textit{4.NBT.A.1}) and descriptions (e.g. \textit{Recognize that in a multi-digit whole number...}) from Achieve the Core's coherence map, which captures how pedagogical experts characterize, organize, and categorize math.\footnote{\prettyUrl{https://github.com/achievethecore/atc-coherence-map/}{github.com/achievethecore/atc-coherence-map}}  
\atc{} includes two types of relationships among standards: as leaves in a topical hierarchy, or as a graph of conceptual connections. Our experiments on verifying (\S\ref{sec:entail}) and tagging (\S\ref{sec:tag}) use these relationships to show how related standards that differ subtly can challenge models. 

As described in \S\ref{sec:task_formats}, the CCSS hierarchy traverses the following levels from top to bottom: grade, $\mathcal{D}$, $\mathcal{C}$, and $\mathcal{S}$.\footnote{In high school, domains analogous to K-8 ones are called ``categories''. More details on how we simplify complexities of CCSS like this one can be found in Appendix~\ref{appdx:tag}.} In addition, \atc{} includes \textit{conceptual connections} between $\mathcal{S}$ that cut across this hierarchy \cite{zimba2018graph}. CCSS is designed to follow students' learning progressions \cite{common2013progressions}, and recognizing these connections is central to enabling students to learn.  
That is, a problem intended to teach students addition would not jump directly to solving for variables. Figure~\ref{fig:v1}, under \textit{Verification}, shows an example of a conceptual connection: the 4th grade $\mathcal{S}$ on the left (\textit{Recognize that in a multi-digit whole number...}) progresses to the 5th grade $\mathcal{S}$ on the right.

\subsection{K-12 Math Problems}\label{sec:problems_activities}

We evaluate models' abilities to identify math skills and concepts using publisher-labeled data pulled from curricular websites. Publicly accessible open educational resources (OER) provide a rich test bed for evaluating models, as they are designed to cover nearly all $\mathcal{S}$ within the grade levels they offer. 
We scrape pre-labeled problems from two OER that have been verified to be reputable by third party curriculum reviewers: Illustrative Mathematics and Fishtank Learning.\footnote{\prettyUrl{https://illustrativemathematics.org/}{illustrativemathematics.org/math-curriculum} and \prettyUrl{https://fishtanklearning.org/}{fishtanklearning.org/about}} Each problem (e.g. the \textit{Task} in Figure~\ref{fig:v1}) in this combined dataset, which we refer to as \mathfish, is a segment of these materials demarcated by $\mathcal{S}$ labels, which we map onto \atc's natural language descriptions. A problem in \mathfish{} can be labeled with multiple $\mathcal{S}$.
Preprocessing details are in Appendix~\ref{sec:data_prep}.

We analyze models in text-only settings, replacing images with a dummy image token, and leave multimodal analyses for future work.
We observed that many examples tend to include tables; we reformat HTML tables into Markdown or JSON based on models' preferences during preliminary experiments (Appendix~\ref{sec:appdx_entail}). Though we contribute this entire scraped dataset of OER problems to facilitate future work, we evaluate models on a smaller evaluation set, which consists of 20\% of all data. We do not evaluate models' problem solving abilities on this data, and instead focus on identifying math skills/concepts, as not all examples contain solutions, and some ``problems'' are designed to be group and/or hands-on activities.

\begin{table}[t]
\centering
\resizebox{\columnwidth}{!}{%
\begin{tabular}{lp{1.5cm}p{1.5cm}p{1.5cm}}
\toprule
\multicolumn{4}{c}{\textbf{\texttt{ATC}: Math Standards Dataset \apple}}                                                                 \\
\midrule
\multicolumn{2}{l}{Grade levels}                        & \multicolumn{2}{l}{K-12} \\
\multicolumn{2}{l}{\# of grades \& domains, e.g. \textit{4.NBT}}    & \multicolumn{2}{l}{65}                             \\
\multicolumn{2}{l}{\# of clusters, e.g. \textit{4.NBT.A}}           & \multicolumn{2}{l}{147}                             \\
\multicolumn{2}{l}{\# of standards, e.g. \textit{4.NBT.A.1}}        & \multicolumn{2}{l}{385}                             \\
\multicolumn{2}{l}{\# of connections b/t $\mathcal{S}$} & \multicolumn{2}{l}{1,040}                             \\ 
\multicolumn{2}{l}{Avg standard description length}        & \multicolumn{2}{l}{36.23}          \\
\midrule
\multicolumn{4}{c}{\textbf{\texttt{MathFish}: Problems Dataset \fish}}                                                         \\ 
\midrule
   & IM  & FL         & Total                 \\ 
\midrule
\# of examples                 & 19,570                        & 2,206         & 21,776                     \\
\# of labeled examples         & 7,735                         & 2,188                        & 9,923                    \\
Total words      & 4.5M                        & 199K                        & 4.7M                     \\
Avg problem length      & 230.51                       & 90.20                       & 216.30                    \\
Grade levels    & K-12            & 3-11             & K-12        \\
\# of unique standards & 366           & 287             & 366       \\
\bottomrule
\end{tabular}
}
\caption{An overview of standards from Achieve the Core (\atc) and curricular materials from \mathfish{}, which combines Illustrative Mathematics (IM) and Fishtank Learning (FL). Lengths and word counts are based on white-spaced tokens. 
}
\label{tbl:data_stats}
\end{table}

\subsection{Problems' Alignment with Standards}

Alignment is not always directly evident in problems' language, and recognizing it requires a deeper understanding of cognitive processes. 
In Figure~\ref{fig:v1}, each $\mathcal{C}$ and $\mathcal{S}$ relate to understanding the base ten number system, but differ in subtle ways. The purpose of the example problem on the left, based on the publisher's description,\footnote{\prettyUrl{https://tasks.illustrativemathematics.org/content-standards/tasks/1562}{tasks.illustrativemathematics.org/content-standards/tasks/1562}} is to teach students how an understanding of the place value system can facilitate division and multiplication by ten. If the problem were instead intended to enable students to \textit{Perform operations with multi-digit whole numbers and with decimals to hundredths}, a wider range of whole numbers than powers of ten would have been included in the problem for students to fully practice those procedural skills. 
As another example of misalignment, the problem in Figure~\ref{fig:v1} does not align with the left $\mathcal{S}$ under \textit{Verification}, as the problem involves place value understanding with decimals, which are not whole numbers. Thus, small differences in mathematical language have distinct consequences for student learning and assessment. 

During evaluation, we assume that all $\mathcal{S}$ not listed in \mathfish{} do not align with a problem. We verify this assumption by having curriculum reviewers label a sample of problems paired with $\mathcal{S}$ that are \textit{not} listed to align with them in publishers' materials, but are \textit{similar} to positive labels. Here, we define ``similar'' as $\mathcal{S}$ that are conceptually connected to positive labels in \atc's map or are their same-$\mathcal{C}$ siblings. We recruited six teachers from a curriculum reviewing organization, which specializes in identifying CCSS alignment of materials. We paid these teachers approximately \$50 an hour, and within an allotted time frame of two weeks, they reviewed 136 pairs of problems with non-listed yet similar $\mathcal{S}$. Only 3 problem and $\mathcal{S}$ pairs in this sample of assumed negatives were judged to actually be positives, thus estimating a ceiling for our assumption.

\section{Verifying Standards Alignment}\label{sec:entail}

This section focuses on our first task format, where models verify the alignment of individual standards against each problem. We first examine how models perform on \mathfish{} problems (\S\ref{sec:verif_exp}), and then apply our best-performing verifier on generated problems (\S\ref{sec:v2}).

\subsection{Can LMs discern (mis)aligned standards?}\label{sec:verif_exp}

\paragraph{Prompt selection.} Since models can be sensitive to prompt design \cite{gonen-etal-2023-demystifying, sclar2023quantifying}, we select prompts from a pool of 15 possible templates based on their performance in small, preliminary experiments (Appendix~\ref{sec:entail_prompts}). These prompts include standards' descriptions to evaluate models' language understanding abilities, and emphasize whether problems teach or enable students to learn a given concept or skill. Few-shot examples are sampled from a small exemplar pool spanning all grade levels (Appendix~\ref{appdx:entail_few_shot}). In the main text, we show results for each model's top performing prompt template, but Appendix~\ref{sec:entail_prompts} elaborates further on the ranges of performance scores we observed. 

\paragraph{Designing problem-standard pairs.} To set up the verification task, we take all gold problem-standard pairs in \mathfish{} to be positive.
For negative labels, we sample them deliberately to vary their closeness to problems' positive ones. 
For each 
problem, we assign five negative $\mathcal{S}$ obtained via different sampling strategies. 
If a problem belongs to grade/s $\mathcal{G}$ and domain/s $\mathcal{D}$, then we sample negative standards from $\mathcal{G}$ or $\mathcal{G'}$ and/or $\mathcal{D}$ or $\mathcal{D'}$. 
Four sampling strategies emerge: ($\mathcal{D}\mathcal{G}$, $\mathcal{D'}\mathcal{G}$, $\mathcal{D}\mathcal{G'}$, $\mathcal{D'}\mathcal{G'}$).
The fifth strategy involves sampling a negative standard that is conceptually connected to, or neighboring~($\mathcal{N}$), positive ones in the \atc{} coherence map. 
We expect that as sampled negatives moves closer to the positive standards in $\mathcal{G}$ and $\mathcal{D}$, the difficulty of discerning true negatives will increase, thus lowering task performance.

\begin{figure}[t]
\centering
\includegraphics[width=\columnwidth]{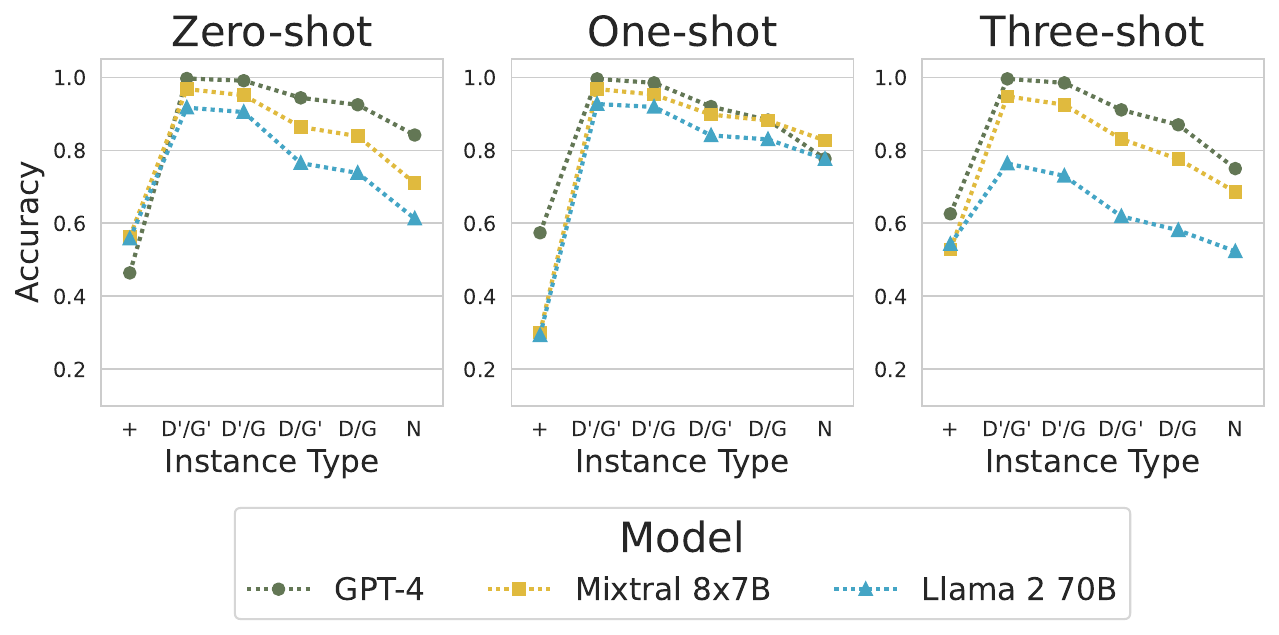}  
\caption{Verification accuracy when problems are paired with aligned standards (+) or with unaligned standards, ordered from left to right in increasing similarity to the positive standard ($\mathcal{D'}\mathcal{G'}$ $\rightarrow$ $\mathcal{D'}\mathcal{G}$ $\rightarrow$ $\mathcal{D}\mathcal{G'}$ $\rightarrow$ $\mathcal{D}\mathcal{G}$ $\rightarrow$ $\mathcal{N}$). Language models have difficulty performing verification as standards become increasingly similar.}
\label{fig:verification_res1}
\end{figure}

\paragraph{Experimental findings.} 
\emph{(i)} As hypothesized, models' accuracy decreases as negative examples become more conceptually similar to positive ones (Figure~\ref{fig:verification_res1}). 
Regardless, the best verifier is three-shot GPT-4, scoring the highest across all positive and negative pairs.
\emph{(ii)} Though \citet{hase2024unreasonable} showed that grade level relates to problem solving performance, our verification task does not have this relationship, as evidenced by insignificant Spearman's $\rho$ between grade level and F1, across all models and prompting approaches (Appendix~\ref{appdx:entail_grade}). 
This suggests that difficult aspects of our task extend beyond how easily a problem can be completed, posing a unique challenge for models. \emph{(iii)} Finally, few-shot exemplars do not uniformly improve performance (Figure~\ref{fig:verification_res1}), nor does pairing task instances with exemplars in the same or nearby grades (Appendix~\ref{appdx:entail_few_shot}).

\begin{table*}[t]
\centering
\resizebox{\textwidth}{!}{%
\begin{tabular}{@{}p{5cm}p{10cm}p{8cm}@{}}
\toprule
\textbf{Error Type} & \textbf{Example $\mathcal{S}$ of a Generated Problem} &  \textbf{Example Teacher Explanation} \\ \midrule
Problem goes beyond the learning stage of the $\mathcal{S}$.  &  1.OA.A.2: \textit{Solve word problems that call for addition of three whole numbers whose sum is less than or equal to 20...}        &   \textit{Total number of beads (11+12) is greater than 20.} [sic] \textit{and therefore beyond the boundary.} \\ \midrule
Problem is too nonsensical or contains incorrect math.  &  G-CO.C.9: \textit{Prove theorems about lines and angles. Theorems include: vertical angles are congruent; when a transversal crosses parallel lines, alternate interior angles are congruent and corresponding angles are congruent...}   &    \textit{Directions are not correct and at best unclear. A transversal line cannot cross 2 parallel lines at 4 points. I think they mean angles, but where the angles are is unclear. And where is angle 3?} \\ \midrule
Problem does not address some part of the $\mathcal{S}$.  &  2.NBT.B.5: \textit{Fluently add and subtract within 100 using strategies based on place value, properties of operations, and/or the relationship between addition and subtraction.}      &   \textit{This is a good problem for practicing addition but it does not include subtraction.}  \\ \midrule
The solution is included as part of the problem setup.   &   7.G.B.4: \textit{Know the formulas for the area and circumference of a circle and use them to solve problems; give an informal derivation of the relationship between the circumference and area of a circle.}  &    \textit{The formulas for area and circumference are provided for students, therefore they are not able to demonstrate that they know the formulas to use independetly} [sic]. \\ \midrule
Problem addresses other concepts/skills. &     HSG-SRT.D.10: \textit{Prove the Laws of Sines and Cosines and use them to solve problems.}  &  \textit{This is not a Law of Sines/Cosines problem. This involved tangent and SOHCAHTOA.}  \\ 
\bottomrule
\end{tabular}
}
\caption{The leftmost column shows common reasons for why generated problems have no or partial alignment, obtained via open coding of teachers' explanations. Provided examples in each row are cases where GPT-4 judges a problem to be fully aligned, but teachers do not. Some $\mathcal{S}$'s descriptions are shortened for brevity.}
\label{tbl:generated_errors}
\end{table*}

\paragraph{Error analysis.} To obtain a closer look at models' errors, we examine the language within particularly challenging $\mathcal{S}$. We find Mixtral and Llama-2's false negatives are related to $\mathcal{S}$ mentioning trigonometry (e.g. \textit{sine}, \textit{cosine}), while GPT-4 does not have this weakness (details in Appendix~\ref{appdx:entail_error}). All three models also struggle with false positives mentioning proportions, ratios, and/or rates. These observations show how a standards-based evaluation framework can help identify model-specific idiosyncrasies in a granular manner. 

We also qualitatively observe misconceptions around the solution strategy targeted by a problem. For example, in one problem, students are asked to \textit{Find the value of each expression mentally. 90-45, 270-45, 270-135, 360-135 ... How did this observation—that the first numbers are all multiples of 90—help you find the value of the differences?}\footnote{\prettyUrl{https://im.kendallhunt.com/k5/teachers/grade-4/unit-7/lesson-2/lesson.html}{im.kendallhunt.com/k5/teachers/grade-4/unit-7/lesson-2/lesson}} In other words, this problem is designed to leverage multiplication to find numerical differences. GPT-4's response when verifying a positive $\mathcal{S}$ related to multiplication instead claims that this problem \textit{focuses on subtraction and mental math strategies... rather than on multiplication of whole numbers..}. This example illustrates how identifying the concepts underlying problems requires recognizing how they are solved.

\subsection{Study 1: Verifying standard alignment of LM-generated problems}\label{sec:v2}

\citet{shapiro2024using}'s survey of teachers showed that some of the most common uses of AI in classrooms include generating materials such as assessments, lesson plans, and assignments. In addition, problem generation is a common task in AI \& education research \citep[e.g.][]{norberg2023rewriting, zhou-etal-2023-learning-analogy,wang-etal-2021-math,mishra2024mathcamps,shah2024aiassisted}, though generations are rarely evaluated by in-domain experts. 
In this section, we investigate the utility of GPT-4, our best verifier from \S\ref{sec:verif_exp}, by applying it onto assessing \textit{generated} math problems, working alongside K-12 math teachers to also obtain their verification judgements.

\paragraph{Generating problems.} To obtain realistic generations, we asked six teachers from \S\ref{sec:problems_activities} to write example queries demonstrating how they would ask a model to generate math problems based on a $\mathcal{S}$. We used teachers' suggested prompts to design 10 prompt templates in which we insert random $\mathcal{S}$ labels and their descriptions to create 100 unique prompts (Appendix~\ref{sec:appdx_v2}). We input these prompts into Llama-2-70B, Mixtral-8x7B, and GPT-4 to generate a total of 300 problems. 

\begin{table}[t]
\centering
\resizebox{0.8\columnwidth}{!}{%
\begin{tabular}{@{}cccc@{}}
\toprule
\textbf{Model} & \textbf{\makecell{Full \\ (GPT-4)}} & \textbf{\makecell{Full \\(teachers)}} & \textbf{\makecell{Full + Partial \\(teachers)}} \\ 
\midrule
Llama-2-70b & 76\% &  19\% & 64\%\\
Mixtral-8x7b & 84\% &   35\%  & 80\% \\
GPT-4 & 96\% &  52\% & 85\% \\
\bottomrule
\end{tabular}
}
\caption{Generated problems' $\mathcal{S}$ alignment, as judged by a GPT-4-based verifier or by teachers.}
\label{tbl:v2_gpt}
\end{table}

\paragraph{Collecting teachers' annotations.}
Sixteen teachers, recruited from the same curriculum reviewing organization as those in \S\ref{sec:problems_activities}, then verified whether these generated problems align with the $\mathcal{S}$ indicated in the original prompt. Teachers were distributed to grade levels based on their prior reviewing and teaching experience. Their annotations include the following labels: ``not a math problem'', ``no alignment'', ``partial alignment'', or ``full-intent alignment'', paired with written explanations. Since these problems are deliberately instructed to align with a $\mathcal{S}$, our generation process leads to many challenging, borderline cases. 
Each problem took 5-15 minutes for a teacher to judge, and we again paid teachers \$50 an hour. We compare teachers' judgements to the best-performing GPT-4 verification approach (\S\ref{sec:verif_exp}). 

\paragraph{Results.} From teachers' judgements in Table~\ref{tbl:v2_gpt}, we see that most generated problems do not fully align with $\mathcal{S}$ indicated in generation prompts. In addition, GPT-4 overestimates problem-standard alignment, and its estimated rates of full-intent alignment even exceed the combined percentage of full and partial ratings from teachers. Thus, this GPT-4 verifier skews more optimistic than teachers as a whole. Still, we observe agreement in the relative ranking of models for problem generation between teachers and our model verifier.
In Table~\ref{tbl:generated_errors}, teachers' written explanations of why a generated problem does not enable a student to learn a given $\mathcal{S}$ reveal the types of pedagogical considerations they make that are missed by problem generators and our GPT-4 verifier. 
Altogether, an LM verifier could be useful for estimating which models may generally generate better aligned problems, but may give less critical judgements than a team of curriculum reviewers.

\section{Tagging Problems with Standards}\label{sec:tag}

As we introduced in \S\ref{sec:data}, tagging navigates a hierarchical decision tree, where each branch presents models a problem and a list of $\mathcal{D}$, $\mathcal{C}$, or $\mathcal{S}$ descriptions. 
We shuffle the ordering of options in each level of the tree to avoid position bias \cite{NEURIPS2023_91f18a12}. The $\mathcal{D}$ level consists of 12 options, while the $\mathcal{C}$ level and $\mathcal{S}$ level each have on average 13.7 options and 2.6 options, respectively. We also give models the option to also respond with ``none'' at each level, indicating that none of the listed options apply to an example problem. Prompt templates and our rationale behind their design can be found in Appendix~\ref{appdx:tag}. In this section, we discuss experiments we run on \mathfish{} problems (\S\ref{sec:tag_results}) and apply our best tagger on GSM8k (\S\ref{sec:v3}).

\subsection{Can LMs traverse the tagging hierarchy?}\label{sec:tag_results}

\paragraph{Prompt selection.} Like with verification, we again run small preliminary experiments using 15 possible prompt templates. In the main text, we present results for the top-performing prompt template for each model.
We also experiment with one-shot and three-shot prompts, using the same problems we used to create few-shot exemplars for verification.  

\paragraph{Findings from assisted traversal setting.} We run several experiments that evaluate models at each level of the decision tree. Rather than have models traverse the tree on their own, we first evaluate in an ``oracle''-assisted setting, where each level is presented with the assumption that the model has correctly chosen the correct branches in the previous level. We compute accuracy per branch based on the fraction of options correctly selected or not selected by the model. For example, if there are five options (\textit{A}, \textit{B}, \textit{C}, \textit{D}, \textit{E}), ground truth is \textit{B}, and the model selects \textit{B} \& \textit{E}, then accuracy on this branch is 4/5.

\begin{figure}[t]
\centering
\includegraphics[width=\columnwidth]{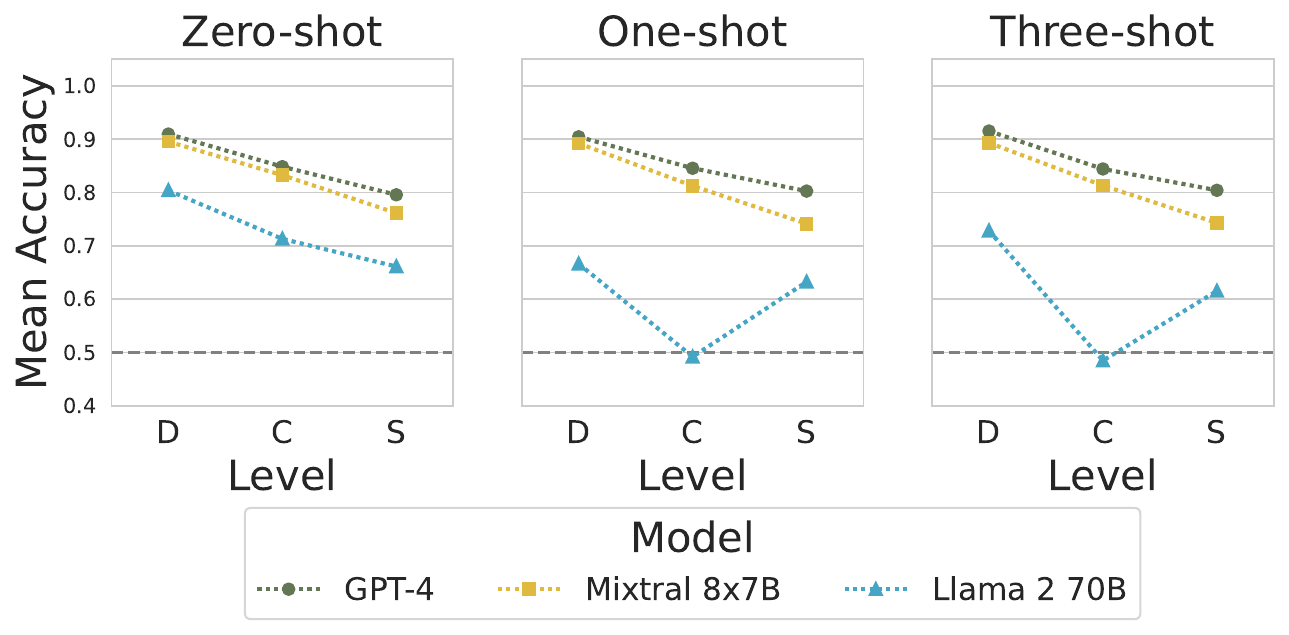}  
\caption{Average per-branch accuracy at each level ($\mathcal{D}$, $\mathcal{C}$, $\mathcal{S}$) of the tagging tree during assisted traversal. The dashed line indicates a random baseline accuracy of 0.5. Stronger models decrease in performance when asked to make more granular decisions.
}
\label{fig:tagging_result}
\end{figure}

Following our hypothesis, stronger LMs, including GPT-4 and Mixtral, tend to perform worse during assisted traversal as levels approach more fine-grained decisions (Figure~\ref{fig:tagging_result}). 
Yet for the weaker Llama-2, few-shot examples can hurt task performance rather than help, which may be due to how some models can struggle with \textit{long} in-context examples more so than others \cite{li2024longcontext}. These model-specific findings also generalize to other prompt templates (Appendix~\ref{appdx:tag}).
Overall, the best performing model and prompt identified from these assisted traversal experiments is three-shot GPT-4 (mean accuracy = 0.850). 

\paragraph{Findings from self-guided traversal setting.} We then run three-shot GPT-4 on a more realistic tagging setup. Here, the model navigates the decision tree by relying on its own decisions in prior levels. Since models are allowed to respond ``none'' at any level of the tree, we are interested in seeing whether our best model and prompt pairing can recover during situations where it traverses down dead ends. We compute self-guided performance at the problem-level via two metrics: weak accuracy, where the model is ``correct'' if predicted $\mathcal{S}$ overlaps with true $\mathcal{S}$, and exact accuracy, where the model is correct only if both sets are equal.

GPT-4 achieves an exact accuracy of only 0.048, and a weak accuracy of 0.502, based on the final predicted set of $\mathcal{S}$ for each problem. In 59.3\% of weakly accurate cases, predicted $\mathcal{S}$ are a superset of gold ones, and GPT-4 tends to assign more $\mathcal{S}$ per problem ($M=3.05$) than gold labels do ($M=1.66$). Exact accuracy varies across gold $\mathcal{D}$, ranging from 0.151 ($\mathcal{D} = $ \texttt{Counting \& Cardinality}) to 0.011 ($\mathcal{D} =$ \texttt{Functions}). When GPT-4 traverses down a dead-end path in earlier levels, it is able to recover by predicting ``none'' at the $\mathcal{S}$ level only 17.4\% of the time. Though these metrics leave much room for improvement, predicted $\mathcal{S}$ are often conceptually near ground truth. They have an average minimum distance of 1.9 edges from true $\mathcal{S}$ in the \atc{} coherence map, which is closer than an average minimum distance of 5.5 edges for random pairs of $\mathcal{S}$. 80.5\% of predicted $\mathcal{S}$ are in the same $\mathcal{D}$ as gold $\mathcal{S}$, and 44.9\% are in the same grade levels. Overall, our best LM setup can \textit{approximate} where problems reside among $\mathcal{S}$, but pinpointing them is still out of reach.

\subsection{Study 2: Tagging problems in GSM8k}\label{sec:v3}

\begin{table}[t]
\centering
\resizebox{\columnwidth}{!}{%
\begin{tabular}{@{}p{10cm}c@{}}
\toprule
\textbf{Most Common $\mathcal{S}$} & \textbf{\#} (\%) \\ 
\midrule
\textbf{4.OA.A.3}. Solve multistep word problems posed with whole numbers and having whole-number answers using the \hlcolor{four operations}, including problems in which remainders must be interpreted... &   250 (20.73\%)             \\
\midrule
\textbf{2.OA.A.1}. Use \hlcolor{addition} and \hlcolor{subtraction} within 100 to solve one- and two-step word problems...  &     200 (16.58\%)   \\ 
\midrule
\textbf{3.OA.D.8}. Solve two-step word problems using the \hlcolor{four operations}. Represent these problems using equations with a letter standing for the unknown quantity...) &   190 (15.75\%)       \\
\bottomrule
\end{tabular}
}
\caption{Excerpts from the top three most common $\mathcal{S}$ tagged on GSM8k's test set. References to $+$ $-$ $\times$ $\div$ are \hlcolor{highlighted}, providing face validity for our tagger, three-shot GPT-4.}
\label{tbl:GSM8k_standards}
\end{table}

Tagging problems with math concepts/skills can help document open math datasets and benchmarks. 
To illustrate this, we apply our task to GSM8k. GSM8k is a popular ``grade school math'' dataset used to evaluate models' problem solving abilities \cite{cobbe2021training}. It has seen uptake by education-related papers \cite{learnlm2024}, as well as in reports benchmarking LMs \citep[e.g.][]{touvron2023llama, chowdhery2023palm}. We expect that most problems in GSM8k should align with $\mathcal{S}$ pertaining to what \citet{cobbe2021training} describe as, ``elementary calculations using basic arithmetic operations ($+$ $-$ $\times$ $\div$)''. Thus, we aim to (1) confirm that GSM8k mostly covers arithmetic and (2) relate a disaggregation of problems by skills/concepts to models' problem solving strengths and weaknesses. 

\paragraph{Experimental setup.} We apply our best performing tagging setup from \S\ref{sec:tag} to GSM8k's test set to approximate what $\mathcal{S}$ it may cover. For math problem solving, we experiment with a range of model sizes within three families: GPT-4, GPT-3.5; Mixtral-8x7b, Mixtral-8x22b; Llama-2-7b, Llama-2-13b, and Llama-2-70b. We run these models on GSM8k's test set using default settings in Eleuther AI's evaluation harness \cite{eval-harness}. 

\begin{figure}[t]
\centering
\includegraphics[width=\columnwidth]{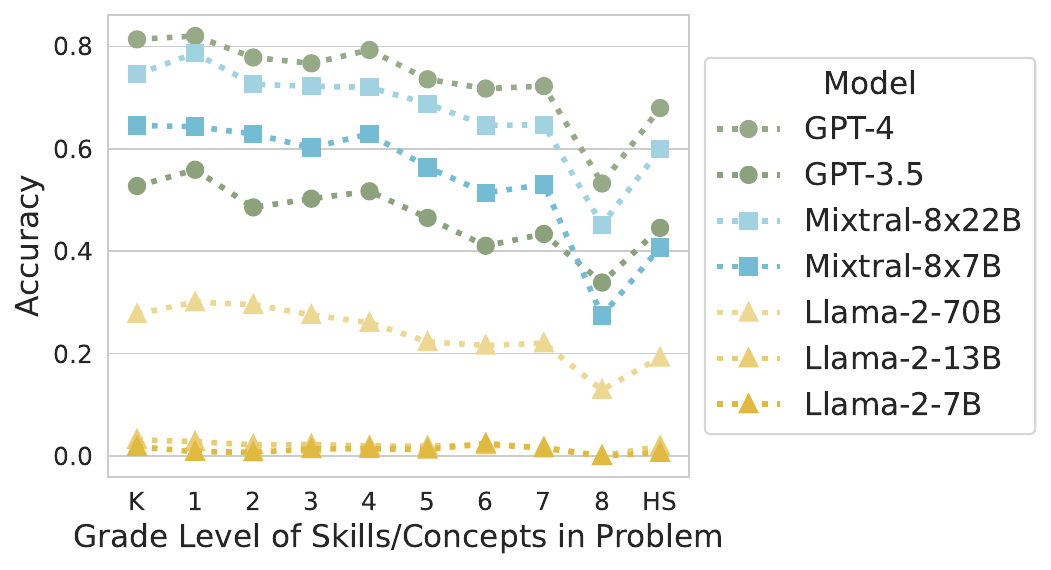}  
\caption{Models' problem solving performance, based on the grade levels of $\mathcal{S}$ tagged in problems.}
\label{fig:v3_model_perf}
\end{figure}

\paragraph{Results} Our tagger estimates that GSM8k covers 32.2\% of all K-12 $\mathcal{S}$ and 56.1\% of K-5 $\mathcal{S}$, with the most common $\mathcal{S}$ related to the four arithmetic operations (Table~\ref{tbl:GSM8k_standards}). So, though GSM8k may be linguistically diverse \cite{cobbe2021training}, it is not necessarily complete nor diverse in its coverage of grade school math skills and concepts. GSM8k is intended to be elementary-level \cite{cobbe2021training}, but some problems land in higher grade levels, e.g. linear equations in 8th grade. The three most frequent $\mathcal{D}$ covered by GSM8k are Operations \& Algebraic Thinking (36.0\%), Expressions \& Equations (25.8\%), and Ratios \& Proportional Relationships (21.8\%). 

Strikingly, across model families and sizes, problem solving performance on GSM8k relates to the grade levels of tagged $\mathcal{S}$ (Figure~\ref{fig:v3_model_perf}). That is, though bigger models overall solve better than smaller ones, they suffer similar rates of performance degradation when they encounter problems involving higher grades' skills and concepts. These trends follow a general intuition that problem solving difficulty should relate to ``hardness'' defined by grade level \cite{hase2024unreasonable}. Thus, even though GSM8k problems do not originally come paired with grade level metadata, a noisy GPT-4 tagger is able to estimate grade-level difficulty using only natural language descriptions of math concepts and skills. 

\section{Conclusion}

Our work investigates how well LMs can identify fine-grained skills and concepts that students learn when completing math problems. We contribute a dataset of 385 expert-designed mathematical standards layered with natural language descriptions and organizational metadata (\atc), a dataset of 9.9K standards-labeled problems drawn from reputable curricula (\mathfish), and annotations of problem and standard pairs by experienced teachers. In addition, we provide tooling to transform data into tasks that assess LMs' abilities to reason about mathematical language and concepts/skills. We find that though LMs still do not reach expert-level performance on verification, tagging, and generation tasks, they show promise and utility for assisting curriculum review and disaggregating math problem solving benchmarks. 

\section{Limitations}\label{sec:limit}

\paragraph{Multimodality.} Math content is inherently multimodal \citep[e.g.][]{lu2023mathvista}, but our tasks focus on text-only data. The problems we gathered not only contain images, but sometimes embed interactive web applets. We plan to include images attached to problems in the final released form of our dataset to facilitate future work. The metadata attached to each example also includes the problem's original url, in case others want to leverage other forms of information present in these online materials. 

\paragraph{Curriculum review.} Throughout the paper, we engage with teachers who professionally review curricular materials. During our process of working with them, we learned the ways in which their annotations for our paper may gloss over the complexities of how curriculum review occurs in practice. For example, when we generate problems based on standards, we include only one standard in the prompt. In reality, teachers may combine multiple standards in a single lesson, or use multiple problems to collectively target one standard. In addition, we asked teachers to individually annotate problem and standard pairs, and their annotations only consist of one pass over these materials (\S\ref{sec:problems_activities}, \S\ref{sec:v2}). During actual reviews, teachers may collect evidence of standards (mis)alignment individually, but later come together for careful deliberation, as flaws observed by one teacher may be missed by another. Finally, curriculum review typically evaluates materials for additional measures of quality beyond CCSS alignment, some of which we briefly discuss in Appendix~\ref{sec:appdx_v2}. We encourage future work to build on these investigations, especially as LMs become increasingly integrated into classrooms and educational technologies \cite{shapiro2024using}.

\section{Ethical Considerations}

LMs show promise as automated tools for gathering and/or suggesting standards (mis)alignment and assisting reviewers in their examination of materials. Though our paper aims to use LMs to automate the task of identifying standards alignment in curriculum, LMs' role in curriculum review and creation processes should be a supporting, rather than leading, one. To design such tools, we believe that it is best to co-create with teachers and curriculum specialists. 

In addition, the curriculum reviewers we worked with may not necessarily reflect the views of all teachers. Though they teach across the U.S., they are predominantly White and female (Appendix~\ref{sec:appdx_v2}). Our work also relies on Common Core, which is established in the U.S. and may not translate to pedagogical standards or practices in other sociocultural contexts.  A range of educators' voices should be forefronted in research that intersects LMs and education.

\atc{} is covered by a Creative Commons Public Domain Dedication License. Within \mathfish, Illustrative Mathematics is licensed as CC BY 4.0, while Fishtank Learning component is licensed under Creative Commons BY-NC-SA 4.0. Both sources are intended to be OER, which is defined as teaching, learning, and research materials that provides users free and perpetual permission to ``retain, reuse, revise, remix, and redistribute'' for educational purposes.\footnote{\prettyUrl{https://guides.library.columbia.edu/OER}{guides.library.columbia.edu/OER}} The transformed versions of these materials as datasets is licensed under ODC-By 1.0, and our code to reproduce our experiments is licensed under Apache 2.0. 

\section{Acknowledgements}

We thank the sixteen teachers from EdReports not only for their annotations, but also for their conversations which greatly informed our work. We also thank curriculum specialists at EdReports for patiently guiding us through the ecosystem of ``real-world'' math curriculum and curriculum review. Our work was generously funded by the Bill \& Melinda Gates Foundation.

\bibliography{custom}

\appendix

\section{Data Collection and Preprocessing}
\label{sec:data_prep}

\subsection{Scraping \mathfish}

\paragraph{Illustrative Math.} Illustrative Math (IM) segments materials across their website based on their intended use in classrooms. We pull problems from the following parts of their website: tasks, lessons, centers, practice, and modeling prompts. $\mathcal{S}$ are listed with varying relations to content, e.g. a problem is ``Building Towards'' a $\mathcal{S}$, if it is foundational but not at the level of a standard. We use the relations ``Addressing'' and ``Alignment'' as ground truth positive labels for $\mathcal{S}$ alignment. The dataset we contribute includes all problems and label types, in case future work wishes to examine discerning how problems that are ``Building On'' a $\mathcal{S}$ may differ from those that are ``Addressing'' or ``Building Towards'' it. 

\paragraph{Fishtank Learning.} We obtain problems from Fishtank Learning (FT) from lessons listed within each unit. FT includes two types of labels: ``Core Standards'' and ``Foundational Standards'', the latter of which is similar to IM's ``Building Towards.'' We use $\mathcal{S}$ listed as ``Core Standards'' as ground truth positive labels for alignment. We again include all problems and label types in the dataset that accompanies our paper. 

\subsection{Label Preprocessing}

\paragraph{Standardization.} $\mathcal{S}$ can be written in a variety of ways by educators and curricular materials (e.g. \textit{HSS-MD.B.5} is the same as \textit{S-MD.5}). We standardize these $\mathcal{S}$ labels so that we can link them across datasets, and use the label version present in Achieve the Core (\atc) as the canonical label. 

\paragraph{Inheritance.} Not all labels present in OER materials are at the $\mathcal{S}$ level. If a $\mathcal{C}$ is listed for a problem, then we infer that the problem aligns with all $\mathcal{S}$ within that $\mathcal{C}$. Similarly, if a sub-standard (e.g. \textit{F-IF.C.7a}, \textit{F-IF.C.7b}) is listed for a problem, we assume it aligns with its parent $\mathcal{S}$. 

\section{Model Wrapper}\label{sec:appdx_model_wrap}

For both verification and tagging, we unify all models under a single model wrapper to keep prompting consistent across them. 
We use the TogetherAI API\footnote{\url{https://docs.together.ai/docs/quickstart}} and OpenAI API\footnote{\url{https://platform.openai.com/docs/overview}} for model access. In cases where a prompt exceeds a model's context window, we truncate the problem description in the prompt, but retain the entirety of $\mathcal{S}$ descriptions. During experiments, all models are given 3 retries for incorrect response formatting (e.g., not including a \textit{yes} or \textit{no} in the verification task format). Retries call the model again with no additional context. Models were run using their default temperature and maximum context window. In total, we spent less than \$5k on API calls.

\section{Verification}\label{sec:appdx_entail}

\subsection{Table Formatting}\label{sec:tables}

\begin{table}[t]
\centering
\resizebox{\columnwidth}{!}{%
\begin{tabular}{@{}lccc@{}}
\toprule
\textbf{Table format} & \multicolumn{1}{l}{\textbf{Mixtral 8x7B}} & \textbf{Llama-2 70B} & \textbf{GPT-4-turbo} \\ \midrule
reStructuredText & 0.855          & 0.666         & 0.882         \\
markdown         & 0.871          & \textbf{0.710} & \textbf{0.880} \\
json             & \textbf{0.885} & 0.696         & 0.869         \\
html             & 0.853          & 0.664         & 0.861         \\ \bottomrule
\end{tabular}
}
\caption{F1 scores during preliminary verification experiments to determine models' table formatting preferences.}
\label{tbl:tables}
\end{table}

Web-scraped math problems sometimes include tables. We first experimented with different table formatting styles in one fixed prompt template: HTML, JSON, Markdown, and reStructuredText. We evaluate on a random sample of 500 verification instances, which consist of OER problems from our evaluation set paired with positive labels or negative labels sampled from $\mathcal{D'}\mathcal{G'}$. Using a fixed prompt template, we find that Llama-2 and GPT-4 prefer Markdown tables, while Mixtral prefers JSON (Table~\ref{tbl:tables}). We retain these table preferences for all further experiments in our paper, including those for tagging. 

\subsection{Prompts}\label{sec:entail_prompts}

Three authors collaboratively wrote a pool of 15 prompt templates $\{ \mathcal{P}_i \mid i = 1, 2, 3, \ldots, 15 \}$ which vary in phrasing. These templates are designed to emphasize whether problems teach or enable students to learn a given concept or skill. Their paraphrases were informed by language occurring in resources that discuss Common Core alignment, especially ``full intent'' or ``focused'' alignment (Figures \ref{fig:entailment_1}-\ref{fig:entailment_15}).\footnote{\url{https://curriculum.illustrativemathematics.org/MS/teachers/design_principles.html}, \url{https://achievethecore.org/page/1118/coherence-map}} We again evaluate on a random sample of 500 examples as we did in \S\ref{sec:tables}. 

Models vary in performance across $\mathcal{P}_i$ during these small preliminary experiments, though there is some overlap across models' top three. Across all 15 prompt templates, we observe F1 ranges of 0.605-0.838 for Llama 2, 0.778-0.895 for Mixtral, and 0.778-0.913 for GPT-4. Each models' top-3 prompts are: $\mathcal{P}_4$ (Figure \ref{fig:entailment_4}), $\mathcal{P}_{10}$ (Figure \ref{fig:entailment_10}), and $\mathcal{P}_{15}$ (Figure \ref{fig:entailment_15}) for Llama 2, $\mathcal{P}_1$ (Figure \ref{fig:entailment_1}), $\mathcal{P}_5$ (Figure \ref{fig:entailment_5}), $\mathcal{P}_{15}$ (Figure \ref{fig:entailment_15}) for Mixtral, and $\mathcal{P}_1$ (Figure \ref{fig:entailment_1}), $\mathcal{P}_4$ (Figure \ref{fig:entailment_4}), $\mathcal{P}_7$ (Figure \ref{fig:entailment_7}) for GPT-4.

\begin{figure}[t]
\centering
\includegraphics[width=\columnwidth]{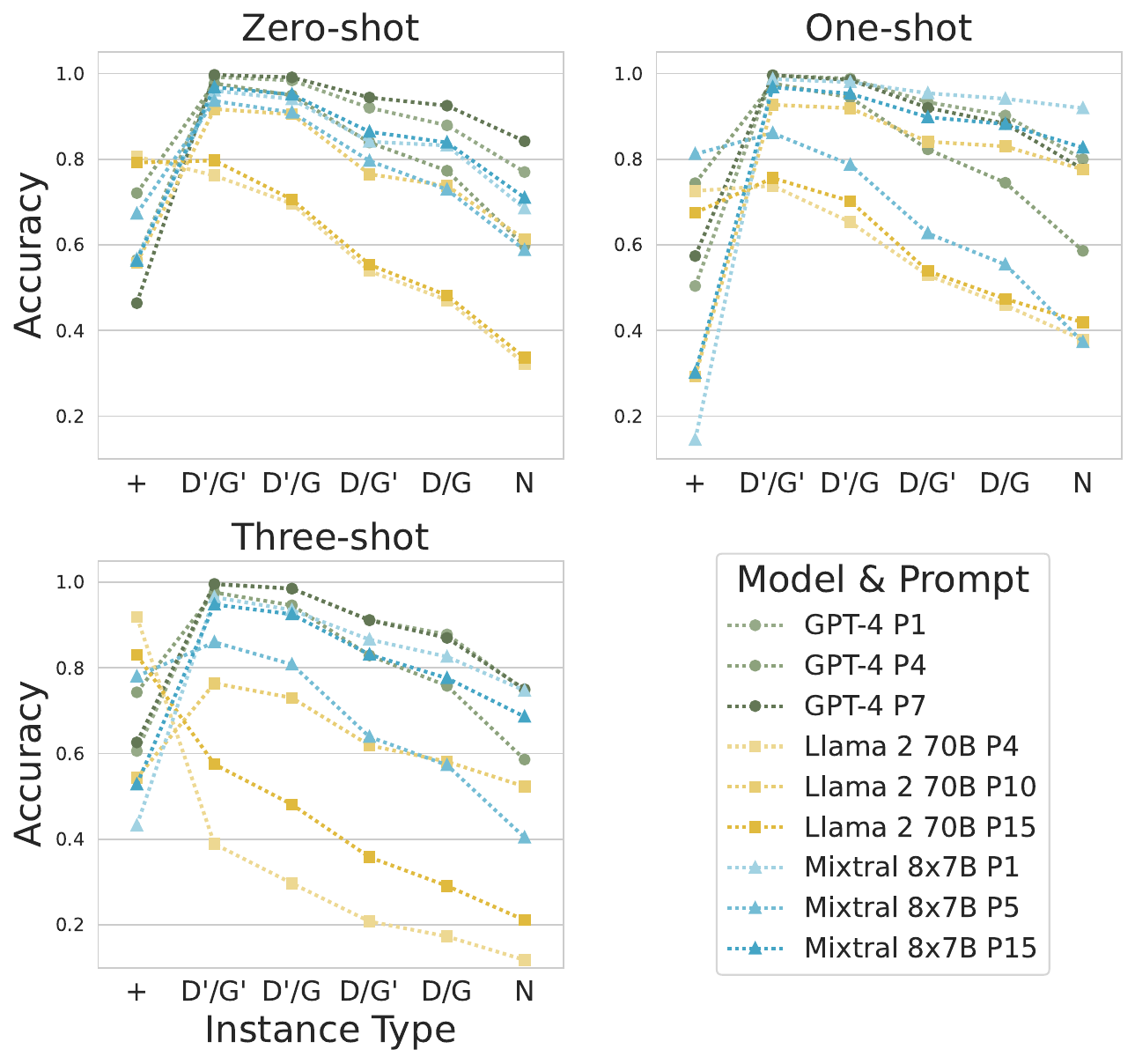}  
\caption{Verification accuracy when problems are paired with aligned standards (+) or with unaligned standards, ordered from left to right in increasing similarity to the positive standard ($\mathcal{D'}\mathcal{G'}$ $\rightarrow$ $\mathcal{D'}\mathcal{G}$ $\rightarrow$ $\mathcal{D}\mathcal{G'}$ $\rightarrow$ $\mathcal{D}\mathcal{G}$ $\rightarrow$ $\mathcal{N}$). Language models have difficulty performing verification as standards become increasingly similar.}
\label{fig:ext_verification_res1}
\end{figure}

Then, we ran these top-3 performing prompt templates on the full evaluation set, across all five types of negative labels as described in the main text \S\ref{sec:entail}. Figure~\ref{fig:ext_verification_res1} shows the variation we obtained in performance across prompts, though general trends reflect the same conclusions as Figure~\ref{fig:verification_res1} in the main text. That is, as negative labels are more similar to positive ones, verification accuracy decreases, and different model and prompt pairings trade off false positives and false negatives differently.

\subsection{Few-Shot Exemplars}\label{appdx:entail_few_shot}

\begin{figure}[t]
\centering
\includegraphics[width=\columnwidth]{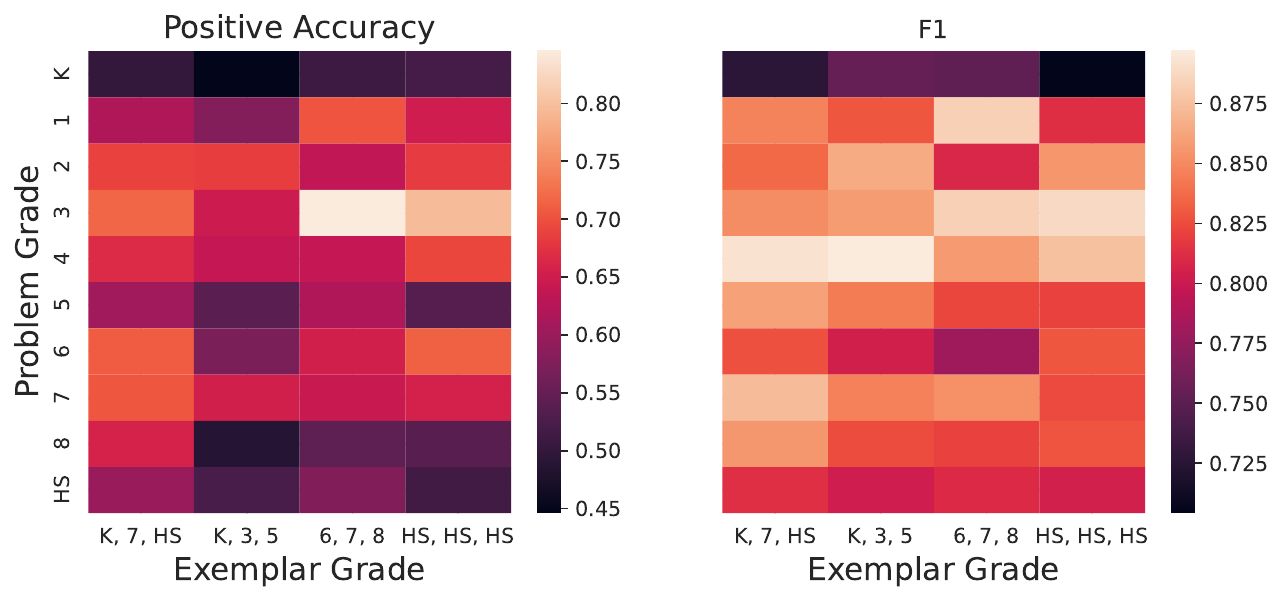}  
\caption{Three-shot verification performance, disaggregated across problems' grade levels and few-shot exemplars' grade levels. We did not find a clear relationship between problem grade and exemplar grade.}
\label{fig:entail_few_shot}
\end{figure}

\begin{figure*}[t]
\centering
\includegraphics[width=\textwidth]{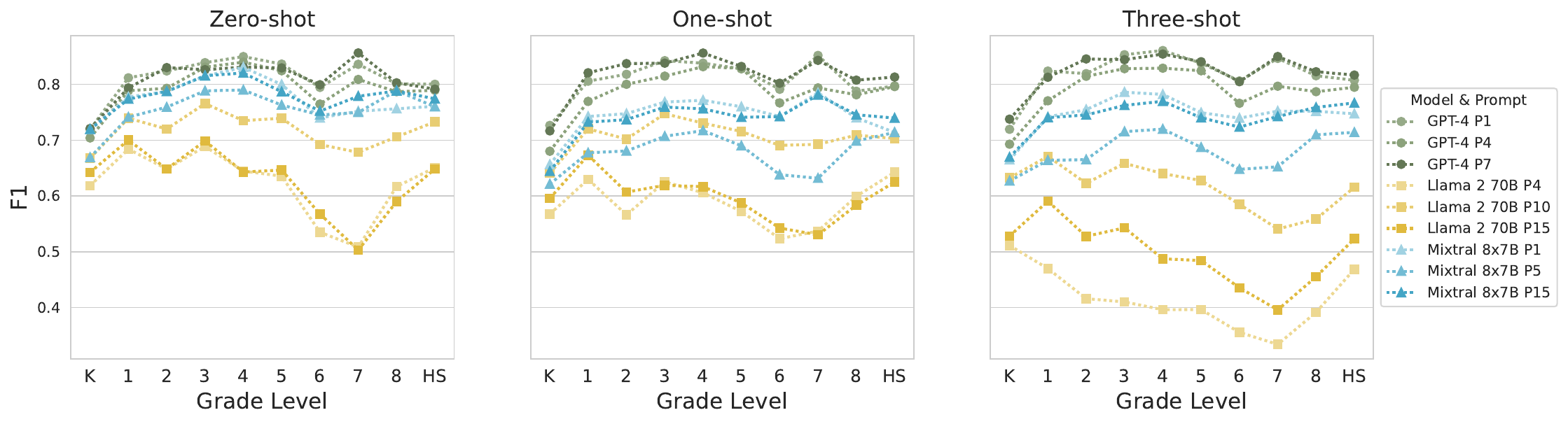}  
\caption{Verification performance across all positive and negative examples, separated by grade level on the $x$-axis. Each model is shown with its top-3 prompt templates, which were identified during preliminary verification experiments.). 
}
\label{fig:grade_entail}
\end{figure*}

We experiment with zero-shot, one-shot, and three-shot prompts. From problems outside of our evaluation set, we pulled one problem from grades K-8 and three from HS. We pair these problems with one positive label and one negative one, which is a randomly selected, conceptually similar neighbor to the problem's positive labels in the \atc{} map. We write Chain-of-Thought-like explanations for these problem and $\mathcal{S}$ pairs to create few-shot exemplars to insert into prompts (e.g. \textit{Example \{i\}} in Figures \ref{fig:entailment_1}-\ref{fig:entailment_15} would be repeated for each exemplar). An example of a few-shot exemplar: \\

\noindent\textbf{Problem:}\\ 
\texttt{\small You’re mixing ingredients for cookies. The recipe says to combine 6 tablespoons, or $\frac{3}{4}$ stick, of butter with 1 cup of sugar. You accidentally mix in a whole stick of butter (8 tablespoons) with the cup of sugar. How can you fix this?}\\

\noindent\textbf{Standard description:}\\
\texttt{\small Use ratio and rate reasoning to solve real-world and mathematical problems, e.g., by reasoning about tables of equivalent ratios, tape diagrams, double number line diagrams, or equations.}\\

\noindent\textbf{Answer:}\\
\texttt{\small yes}\\

\noindent\textbf{Explanation:}\\
\texttt{\small This problem solves a real-world problem involving mixing ingredients for cookies. A student doing this problem would need to reason about equivalent ratios of butter amounts to sugar amounts.}\\

For one-shot prompts, we randomly select an exemplar problem and $\mathcal{S}$ pair from any grade. For three-shot prompts, we select exemplars that span wide and narrow grade ranges: \{(K, 7, HS), (K, 3, 5), (6, 7, 8), (HS, HS, HS)\}. We initially hypothesized that models may perform better on problems accompanied by few-shot exemplars in similar grade spans, but we did not confirm this hypothesis with our exemplar pool (Figure~\ref{fig:entail_few_shot}). 

\subsection{Error Analysis}\label{appdx:entail_error}

In the main paper, we discuss false negative and false positive patterns among models by referring to the language within particularly challenging $\mathcal{S}$. To do this analysis, we tokenize $\mathcal{S}$ using \texttt{Bling Fire}, and remove tokens that are less than 2 characters long, are \texttt{nltk} English stopwords, or appear in less than 5 $\mathcal{S}$. Table~\ref{tbl:error_words} shows the results of this analysis. 

\subsection{Performance Across Grade Levels}\label{appdx:entail_grade}

Intuitively, higher grade levels may suggest lower verification performance, based on known measures of problem ``hardness'' \cite{hase2024unreasonable}. However, Figure~\ref{fig:grade_entail} shows a lack of a consistently positive trend between F1 (calculated over all positive and negative examples) and problems' grade level. For all 27 combinations of model and prompting approaches, we did not observe a significantly positive Spearman $\rho$ between performance and grade level, where significance is measured as $p < 0.05$ with Bonferroni correction.

\begin{table}[t]
\centering
\resizebox{\columnwidth}{!}{%
\begin{tabular}{@{}lp{11cm}@{}}
\toprule
\textbf{Error} & \textbf{Words in $\mathcal{S}$ (Error Rate)} \\ 
\midrule
\multicolumn{2}{l}{Model: three-shot \textbf{GPT-4}}  \\ 
\midrule
FP & proport (0.26), ratio (0.24), rate (0.24), input (0.22), verbal (0.2), quantiti (0.2), descript (0.19), relationship (0.19), assign (0.18), equival (0.18)\\
FN & invers (0.74), argument (0.69), half (0.68), unlik (0.61), partit (0.61), similar (0.6), congruenc (0.6), prove (0.6), cylind (0.58), first (0.56)\\
\midrule
\multicolumn{2}{l}{Model: one-shot \textbf{Mixtral}}  \\  
\midrule
FP & proport (0.27), extend (0.26), previou (0.25), featur (0.25), reason (0.22), mathemat (0.22), assess (0.21), quantiti (0.21), diagram (0.2), neg (0.2)\\
FN & sine (1.0), cosin (1.0), polynomi (0.97), invers (0.95), half (0.91), congruent (0.91), similar (0.9), definit (0.89), transform (0.88), argument (0.87)\\
\midrule
\multicolumn{2}{l}{Model: one-shot \textbf{Llama-2}}  \\ 
\midrule
FP & proport (0.35), person (0.28), rate (0.28), extend (0.27), previou (0.27), ratio (0.26), per (0.25), quantiti (0.25), hour (0.24), descript (0.24)\\
FN & ident (1.0), sine (0.91), cosin (0.91), name (0.89), count (0.88), true (0.88), 1,000,000 (0.88), need (0.88), partit (0.87), origin (0.85)\\
\bottomrule
\end{tabular}
}
\caption{Stemmed words in $\mathcal{S}$ that are most difficult for each model's best prompting approach, with error rates in parentheses. FN = false negative, FP = false positive.}
\label{tbl:error_words}
\end{table}

\section{Tagging}\label{appdx:tag}

\subsection{Prompts}\label{sec:tag_prompts}

We write 15 possible prompt templates $\{ \mathcal{Q}_i \mid i = 1, 2, 3, \ldots, 15 \}$ and run models on 500 random instances of a simple toy task to filter out catastrophic templates. In this toy task, we provide models an OER problem and 5 random $\mathcal{S}$ descriptions, and models are asked to select positive labels hidden among these options. 

Several elements in tagging prompt templates (indicated in curly brackets in Figures~\ref{fig:tagging_1}-\ref{fig:tagging_12}) vary depending on the level of the tagging decision tree. We outline these elements here: \\
\\
Domain
\vspace{-3mm}
\begin{itemize}
\itemsep-0.5em 
\item \texttt{relation\_definition} = `'
\item \texttt{level} = `topics'
\item \texttt{Level} = `Topic'
\item \texttt{relation} = `teaches'
\item \texttt{options} = These map onto K-8 domains and HS categories, but are not a one-to-one mapping. Some high school (HS) categories are equivalent or similar to a domain in K-8, and some differences in K-8 domains are difficult to explain a brief description at the domain-level. Thus, a ``domain'' in our paper sometimes groups multiple actual CCSS domains/categories. We mostly retain the original CCSS K-8 domains and HS categories, but make exceptions for the following: we group OA (Operations \& Algebraic Thinking), EE (Expressions \& Equations), and A (HS Algebra) into \textit{Operations \& Algebra}, S (HS Statistics \& Probability) and SP (K-8 Statistics \& Probability) to \textit{Statistics \& Probability}, and finally NS (K-8 The Number System) and N (HS Number and Quantity) to \textit{Number Systems and Quantity}. Since CCSS and \atc{} do not provide brief descriptions of domains, we worked with a curriculum specialist to write these descriptions of each $\mathcal{D}$ option (Figure~\ref{fig:domain_descriptions}). 
\end{itemize}
Cluster
\vspace{-3mm}
\begin{itemize}
\itemsep-0.5em 
\item \texttt{relation\_definition} = `'
\item \texttt{level} = `mathematical concepts/skills'
\item \texttt{Level} = `Mathematical concepts/skill'
\item \texttt{relation} = `teaches'
\item \texttt{options} = Options are natural language descriptions of clusters, obtained from \atc.
\end{itemize}
Standard
\vspace{-3mm}
\begin{itemize}
\itemsep-0.5em 
\item \texttt{relation\_definition} = `A problem or activity aligns with a standard if it can enable students to learn the full intent of the concepts and skills outlined in the standard's description.'
\item \texttt{level} = `standards'
\item \texttt{Level} = `Standard'
\item \texttt{relation} = `aligns with'
\item \texttt{options} = Options are natural language descriptions of standards, obtained from \atc.
\end{itemize}

We observed models struggling to follow response formatting instructions in early experiments. Thus, our prompt paraphrases also vary specifications around response format, e.g. comma-separated selected options (Figure~\ref{fig:tagging_1}). In addition, $\mathcal{P}_7$-$\mathcal{P}_{15}$ suggested walking through the steps of solving or completing a given problem (e.g. Figure~\ref{fig:tagging_11}). To rank $\mathcal{Q}_i$ per model, we first calculate the rate to which models format responses correctly, and break ties based on models' weak accuracy, where responses are correct if predictions overlap with gold labels.

Models, like with verification, vary in performance across all 15 $\mathcal{Q}_i$. On our toy task, we observe exact accuracy ranges of 0.046-0.332 for Llama 2, 0.304-0.542 for Mixtral, and 0.586-0.75 for GPT-4. This performance ordering of models parallels the one obtained for verification, though the gaps among models and prompts are greater here. Some $\mathcal{Q}_i$ favored by one model are highly detrimental to another, and asking a model to walk through doing a problem does not necessarily improve performance. Each models' top-3 prompts are: $\mathcal{Q}_3$ (Figure~\ref{fig:tagging_3}), $\mathcal{Q}_{11}$ (Figure~\ref{fig:tagging_11}), $\mathcal{Q}_{12}$ (Figure~\ref{fig:tagging_12}) for Llama 2, $\mathcal{Q}_5$ (Figure~\ref{fig:tagging_5}), $\mathcal{Q}_{6}$ (Figure~\ref{fig:tagging_6}), $\mathcal{Q}_{12}$ (Figure~\ref{fig:tagging_12}) for Mixtral, and $\mathcal{Q}_1$ (Figure~\ref{fig:tagging_1}), $\mathcal{Q}_{4}$ (Figure~\ref{fig:tagging_4}), $\mathcal{Q}_{10}$ (Figure~\ref{fig:tagging_10}) for GPT-4. 

\begin{figure}[t]
\centering
\includegraphics[width=\columnwidth]{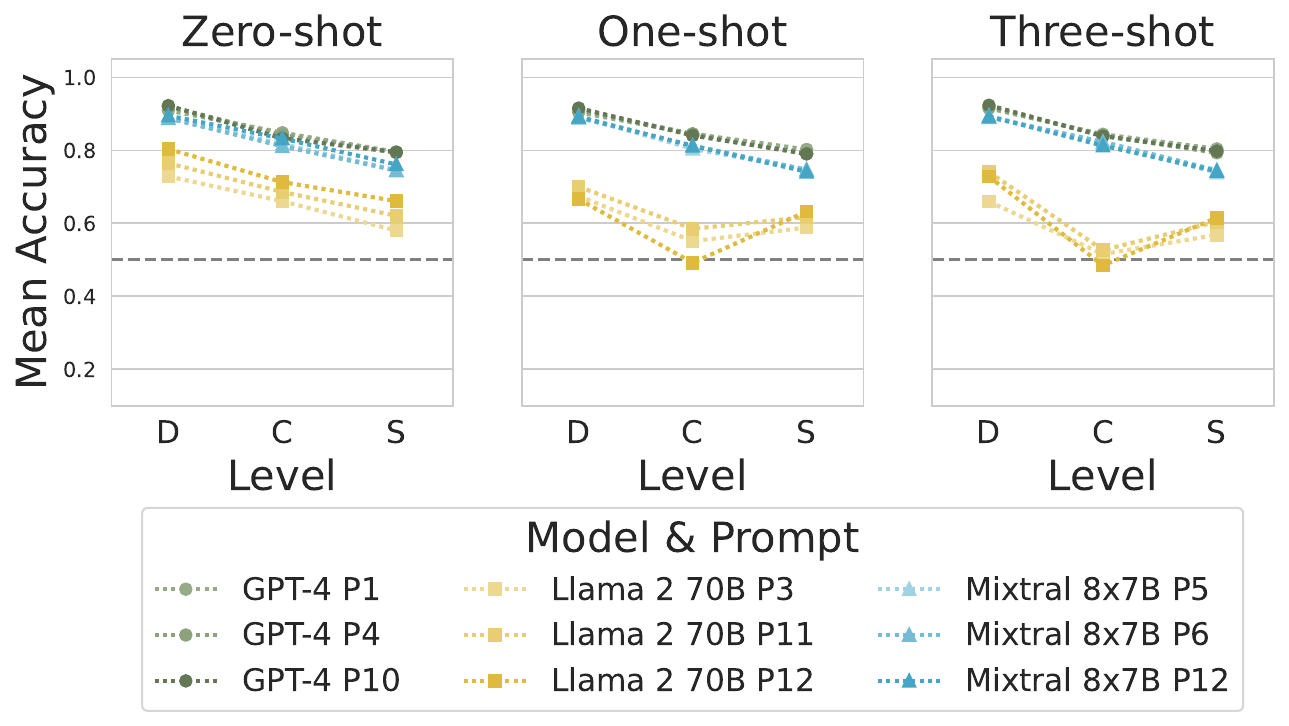}  
\caption{Average per-branch accuracy at each level ($\mathcal{D}$, $\mathcal{C}$, $\mathcal{S}$) of the tagging tree during assisted traversal. The dashed line indicates a random baseline accuracy of 0.5. Stronger models decrease in performance when asked to make more granular decisions.
}
\label{fig:ext_tagging_result}
\end{figure}

We ran these top-3 performing prompt templates for each model on the full evaluation set, and find that the trends we describe in the main text's Figure~\ref{fig:tagging_result} generalize to other top-performing prompts (Figure~\ref{fig:ext_tagging_result}). That is, as stronger models are asked to make more granular decisions from $\mathcal{D}$ to $\mathcal{C}$ to $\mathcal{S}$, their performance decreases. 

\subsection{Few-Shot Exemplars}\label{sec:appdx_tag_few}

We again wrote explanations for $\mathcal{D}$-, $\mathcal{C}$-, and $\mathcal{S}$-level tagging decisions to create exemplars, and sampled one-shot and three-shot exemplars in the same manner as we did with verification. An example of an exemplar at the $\mathcal{S}$-level: \\

\noindent\textbf{Problem:}\\ 
\texttt{\small Task\\
Cruz and Erica were both getting ready for soccer.\\
Cruz ran 1 lap around the school.\\
Erica ran 3 laps around the playground.\\
Erica said,\\
I ran more laps, so I ran farther.\\
Cruz said,\\
4 laps around the school is 1 mile, but it takes 12 laps around the playground to go 1 mile. My laps are much longer, so I ran farther.\\
Who is right? Draw a picture to help you explain your answer.}\\

\noindent\textbf{Options:}\\
A. \texttt{\small Explain why a fraction a/b is equivalent to a fraction (n × a)/(n × b) by using visual fraction models, with attention to how the number and size of the parts differ even though the two fractions themselves are the same size. Use this principle to recognize and generate equivalent fractions. Grade 4 expectations in this domain are limited to fractions with denominators 2, 3, 4, 5, 6, 8, 10, 12, and 100.} \\
B. \texttt{\small Compare two fractions with different numerators and different denominators, e.g., by creating common denominators or numerators, or by comparing to a benchmark fraction such as 1/2. Recognize that comparisons are valid only when the two fractions refer to the same whole. Record the results of comparisons with symbols >, =, or <, and justify the conclusions, e.g., by using a visual fraction model. Grade 4 expectations in this domain are limited to fractions with denominators 2, 3, 4, 5, 6, 8, 10, 12, and 100.}\\

\noindent\textbf{Answer:}\\
\texttt{\small A, B}\\

\noindent\textbf{Explanation:}\\
\texttt{\small This task asks students to compare two fractions that are equal. The two fractions have different numerators and denominators (e.g. 1/4 and 3/12). The students need to explain why the two fractions are equal by drawing a picture, which involves using visual fraction models.}\\

\section{Generated Problems Study}\label{sec:appdx_v2}

\begin{table}[t]
\centering
\resizebox{\columnwidth}{!}{%
\begin{tabular}{@{}llrr@{}}
\toprule
\textbf{} & \textbf{} & \multicolumn{1}{c}{\textbf{Count}} & \multicolumn{1}{c}{\textbf{Percentage}} \\ \midrule
\multirow{2}{*}{\textbf{Race}}             & Black/African American & 2  & 12.50\% \\
                                           & White (Caucasian)      & 14 & 87.50\% \\
\midrule
\multirow{3}{*}{\textbf{Gender}}           & Female                 & 13 & 81.25\% \\
                                           & Male                   & 1  & 6.25\%  \\
                                           & Prefer not to answer   & 2  & 12.50\% \\
\midrule
\multirow{4}{*}{\textbf{Region}}           & West                   & 2  & 12.50\% \\
                                           & Southeast              & 3  & 18.75\% \\
                                           & Northeast              & 4  & 25.00\% \\
                                           & Midwest                & 7  & 43.75\% \\
\midrule
\multirow{4}{*}{\textbf{Yrs in Education}} & 0-10 years             & 4  & 25.00\% \\
                                           & 11-20 years            & 3  & 18.75\% \\
                                           & 21-30 years            & 7  & 43.75\% \\
                                           & 30+ years              & 2  & 12.50\% \\ \bottomrule
\end{tabular}
}
\caption{Teachers' experiential and sociodemographic backgrounds.}
\label{tbl:demographics}
\end{table}

We worked with sixteen U.S.-based, K-12 teachers from a curriculum reviewing organization to annotate pairs of problems and standards (\S\ref{sec:problems_activities}, \S\ref{sec:v2}). These teachers differ from teachers one may recruit from other sources (e.g. Upwork) in that they have prior experience reviewing Common Core alignment of published curricular materials. We worked with three cohorts of 4-6 of teachers, through face-to-face virtual meetings to provide an initial set of instructions, and then communicated with them throughout the annotation process to clarify questions. These teachers were explicitly told that their annotations and explanations, but no identifying information, would be released as part of this project. Each teacher was paid a stipend that averaged \$50 an hour. Most of these teachers are White women, and they span regions across the U.S. and have varying lengths of experience in education (Table~\ref{tbl:demographics}). 

\subsection{Creating Prompt Templates}

\begin{table}[t]
\centering
\resizebox{\columnwidth}{!}{%
\begin{tabular}{@{}p{12cm}@{}}
\toprule
\textbf{Prompt Templates for Generating Problems}                                                                                 \\ \midrule
Create a problem for \{standard\}. This problem will be for a special education resource student.                                 \\\midrule
Create a math problem for \{standard\}. This problem will need to be simplified to support diverse learners in the resource room. \\\midrule
Create a math problem that uses \{standard\}.                                                                                     \\\midrule
please generate a possible quiz item aligned to common core SS for Math for \{standard\}                                          \\\midrule
Please generate an item aligned to CCSS math \{standard\} for students to use to practice with.                                   \\\midrule
my math class is very diverse, with many nationalities, races, languages, and genders represented along with diverse familiy structures.  Please generate a culturally competent practice item for use in an assessment with the standard \{standard\}. \\\midrule
Please generate a practice problem aligned to \{standard\}.                                                                       \\\midrule
Create a multi-step word problem aligned to \{standard\} with at least 3 steps to solve. Make the the problem engaging and relevant for kids who are interested in culturally responsive, real-life scenarios that are fun. Make it easily adjustible for me to change some words/numbers around or edit as needed. \\\midrule
Can you create a problem aligned to \{standard\} that requires students conceptually understand this standard? Make one that is enaging, open, and culturally responsive. \\\midrule
create a problem aligned to \{standard\} in kid friendly language or broken down for ELL students to understand.                  \\ \bottomrule
\end{tabular}
}
\caption{Prompt templates inspired by teachers' suggestions, which we used for generating math problems in \S\ref{sec:v2}.}
\label{tbl:generation_prompts}
\end{table}

First, we asked a subset of teachers (our first cohort) to write prompts that reflect the following: \textit{Pretend you have 1-3 standards (not the MP standards) in mind that you want to teach. How would you ask a model to generate a math problem based on these standard/s?} Here, ``MP standards'' refer to mathematical practice standards, while our work focuses on CCSS math \textit{content} standards. 

We found that teachers usually refer to standards by their labels, e.g. \textit{4.NBT.A.1}, when searching for curricular resources, rather than specify full standard descriptions. In addition, they do not always specify in their prompts that the standards are CCSS standards, rather than some other set of standards that may use similar labels (e.g. state-specific standards). Thus, at the end of each prompt template, we appended each CCSS standard's natural language description to better guide models, e.g. \texttt{Standard \{label\}: \{description\}.}

Altogether, teachers wrote around 20 prompts, though due to variation in teachers' prior experience with LMs, some prompts were not suitable as instruction-like inputs, e.g. \textit{I'd like to see if AI could quickly generate sets of problems with specific root types, such as integers, fractions, etc both with the coefficient of $x^2$ as 1 and with it as an integer.} Thus, we only select only a subset of all proposed prompts, especially ones that could be minimially edited into prompt templates (Table~\ref{tbl:generation_prompts}). We remove extraneous, standard-specific information, e.g. \textit{My students need to prove they can add and subtract within 10 using models}, to make prompts generalizable for nearly all standards. Teachers' suggested prompts varied in complexity, including simple requests such as \textit{Create a math problem that uses \{$\mathcal{S}$\}} to ones that include details around their students' needs, e.g. \textit{This problem will need to be simplified to support diverse learners in the resource room}. The diversity within teachers' suggested prompts, though, could be informative for future work that extends the evaluation of problem generation beyond template-based inputs. 

We observed that some teachers wished to teach multiple standards within a single generated problem, but to scope this study, we simplify all prompt templates to include only a single standard. Thus, to make our setup reflective of standards that may be realistically taught in isolation within a problem, when sampling standards to insert into prompts, we chose ones that appear in \mathfish{} with singly-labeled problems. 

\subsection{Other Observations of Generated Problems}

Aside from standards alignment issues, teachers wrote down additional observations that pertain to the quality of generated problems. We outline two common ones here, in case they are informative for guiding more extensive future work. 

\paragraph{Readability.} Teachers commented on the readability of generated problems. Our prompts do not explicitly indicate the audience of these generations, and generally, these problems could serve two overarching purposes: materials for teachers to work through with students, or materials placed directly in front of students to work on independently. For the former case, some teachers, especially those annotating high school problems, noted that LaTeX formatting outputted by models was difficult for them to parse. For the latter case, some generated problems, especially those addressing lower grades' standards, were not suitable for those students' reading levels. 

\paragraph{Cultural competency.} Some teachers also commented that LMs' attempt to produce culturally competent or culturally responsive practice items was only done at the surface level, e.g. activities that involve ``celebrating diversity,'' without deeper engagement with established frameworks around culturally responsive pedagogy \citep[e.g.][]{bonner2021practicing,ladsonbillings2021}. In addition, it's possible that teachers' prompts were underspecified, and to produce actually culturally aligned problems, teachers needed to explain the context in which they are teaching. For example, one teacher wrote, \textit{culturally responsive leads to an African village? Doesn't seem truly culturally relevant to most students in the US.} Local context also matters, as some topics are prohibited in some schools. For example, one teacher commented that one generated problem mentioning same-sex marriage and non-binary people is not legally allowed to be taught in public schools in their state.

Finally, we also observe cases where generations from different models contain eerily similar wording. For example, Llama-2 and Mixtral both generated problems containing \textit{A bag contains 5 red balls, 7 blue balls, and 3 green balls.} This can suggest memorization of pretraining or finetuning data, and implies a lack of linguistic or topical diversity among problems generated across models. 

\begin{figure*}[h]
    \centering %
    \begin{tcolorbox}[
    otherText,
    title={\textbf{Domain descriptions}},
    ]
\small
- Counting \& Cardinality: students learn to know number names and the count sequence, count to tell the number of objects, and compare numbers.\\
- Operations \& Algebra: students learn to solve problems using algebraic thinking and operations such as addition, subtraction, multiplication, and division. They may learn to identify and explain arithmetic patterns, evaluate or manipulate numerical expressions, and reason with equations and inequalities. \\
- Number \& Operations in Base Ten: students learn to work with the base-ten system and build place value understanding. \\
- Measurement \& Data: students learn to work with data and measure attributes such as time, money, length, area, and volume. They may learn to compare measurements to operations and convert between different units of measure.\\
- Geometry: students learn to classify geometric figures by their properties, and understand and compare the relationships between them. Students develop and use formulas to compute lengths, areas, and volumes, and they use transformations to generate new shapes and compare existing ones. Geometry can be studied with and without coordinates.\\
- Number \& Operations - Fractions: students learn to understand fractions as numbers and may work with them using addition, subtraction, multiplication, and division.\\
- Ratios \& Proportional Relationships: students learn to recognize, describe, represent, and reason with ratios, rates, proportional relationships, and percent. \\
- Number Systems and Quantity: students learn to understand the complex number system and reason quantitatively.\\
- Statistics \& Probability: students learn to analyze and produce data distributions, and build understanding of univariate and bivariate data. They learn to interpret data, make statistical inferences, justify conclusions, understand rules of probability, and use probability to make decisions. \\
- Functions: students learn to define, use, and evaluate functions to model relationships between quantities. \\
- Modeling: students learn to choose and use appropriate mathematics and statistics to analyze real-world empirical situations, improve decisions, and report on their conclusions and the reasoning behind them.

    \end{tcolorbox}
    \caption{Brief descriptions we wrote to include as options for $\mathcal{D}$-level tagging prompts.}
    \label{fig:domain_descriptions}
\end{figure*}

\begin{figure*}[h]
    \centering %
    \begin{tcolorbox}[
    entailmentPrompt,
    title={\textbf{Prompt 1 for verification task}},
    ]
    \small
You are a math expert reviewing K-12 curricular materials. You will be shown a problem or activity obtained from school curriculum and a description of math content. Your task is to assess whether the problem or activity aligns with the provided description. Answer `yes' if it does align, and `no' it does not. \\

Example \{i\}: \\
Problem/activity: \\
\{example problem activity\} \\

Description: \\
\{example standard description\} \\

Answer: \\
\{example answer\} \{example thought\} \\

Now, assess whether the following problem or activity aligns with the provided description. \\

Problem/activity: \\
\{problem activity\} \\

Description: \\
\{standard description\}
    \end{tcolorbox}
    \caption{Prompt 1 for verification.   
    \label{fig:entailment_1}}
\end{figure*}

\begin{figure*}[h]
    \centering %
    \begin{tcolorbox}[
    entailmentPrompt,
    title={\textbf{Prompt 4 for verification task}},
    ]
\small
You are a math instructor reviewing problems and activities meant to support K-12 students in learning mathematical skills and concepts. You will be shown a problem or activity obtained from school curriculum and a description of mathematical concepts and skills. Your task is to determine whether the problem or activity can enable students to understand the concepts or skills in the provided description. Answer `yes' if it does, and `no' if it does not. \\

Example  \{i \}:\\
Concept/skill: \\
 \{example standard description\}\\

Problem/activity:\\
 \{example problem activity\}\\

Answer: \\
 \{example answer\}  \{example thought\}\\

Now, determine whether the following problem or activity can enable students to understand the concepts or skills in the provided description.\\

Concept/skill: \\
 \{standard description\}\\

Problem/activity: \\
 \{problem activity\}\\

    \end{tcolorbox}
    \caption{Prompt 4 for verification.   
    \label{fig:entailment_4}}
\end{figure*}

\begin{figure*}[h]
    \centering %
    \begin{tcolorbox}[
    entailmentPrompt,
    title={\textbf{Prompt 5 for verification task}},
    ]
\small
You are a skilled math instructor and curriculum specialist for K-12 mathematics, specifically the Common Core. Your job is to assess whether a problem or activity is `aligned' with a given Common Core standard in Mathematics. A problem is `aligned' with a standard if the problem or activity helps students fully understand or learn the concept or skill described in the standard. If the problem or activity only helps students learn part but not all of a standard, then it does not align. Answer 'yes' if the problem aligns with the standard or `no' if not. \\ 

Example \{i\}:\\
Concept/skill: \\
\{example standard description\}\\

Problem/activity:\\
\{example problem activity\}\\

Answer: \\
\{example answer\} \{example thought\}\\

Now, assess whether the following problem or activity is 'aligned' with the given Common Core standard in Mathematics.\\

Concept/skill: \\
\{standard description\}\\

Problem/activity: \\
\{problem activity\}\\

    \end{tcolorbox}
    \caption{Prompt 5 for verification.   
    \label{fig:entailment_5}}
\end{figure*}

\begin{figure*}[h]
    \centering %
    \begin{tcolorbox}[
    entailmentPrompt,
    title={\textbf{Prompt 7 for verification task}},
    ]
\small
You are a skilled math instructor and curriculum specialist for K-12 mathematics, specifically the Common Core. Your job is to assess whether a problem or activity is `aligned' with a given Common Core standard in Mathematics. A problem is `aligned' with a standard if the problem or activity helps students fully understand or learn the concept or skill described in the standard. If the problem or activity only helps students learn part but not all of a standard, then it does not align. Answer `yes' if the problem aligns with the standard or `no' if not.   \\

Example \{i\}:\\
Problem/activity:\\
\{example problem activity\}\\

Concept/skill: \\
\{example standard description\}\\

Answer: \\
\{example answer\} \{example thought\}\\

Now, assess whether the following problem or activity is 'aligned' with the given Common Core standard in Mathematics.\\

Problem/activity: \\
\{problem activity\}\\

Concept/skill: \\
\{standard description\}\\

    \end{tcolorbox}
    \caption{Prompt 7 for verification.   
    \label{fig:entailment_7}}
\end{figure*}

\begin{figure*}[h]
    \centering %
    \begin{tcolorbox}[
    entailmentPrompt,
    title={\textbf{Prompt 10 for verification task}},
    ]
\small
You are a math expert reviewing K-12 curriculum. Does this problem or activity enable students to completely learn the following concept or skill? Answer `yes' if it does, and `no' if it does not. Answer no if the problem helps students understand part but not all of a concept or skill. \\

Example \{i\}: \\
Problem/activity: \\
\{example problem activity\} \\

Concept/skill:  \\
\{example standard description\} \\

Answer:  \\
\{example answer\} \{example thought\} \\

Now, does the following problem or activity enable students to learn the full intent of the following concept or skill? \\

Problem/activity:  \\
\{problem activity\} \\

Concept/skill:  \\
\{standard description\} \\
    \end{tcolorbox}
    \caption{Prompt 10 for verification.   
    \label{fig:entailment_10}}
\end{figure*}

\begin{figure*}[h]
    \centering %
    \begin{tcolorbox}[
    entailmentPrompt,
    title={\textbf{Prompt 15 for verification task}},
    ]
\small
You are a math expert reviewing K-12 curriculum to assess whether it addresses specific mathematical standards. Does the problem or activity shown below enable students to learn the full intent of the following concept or skill? Answer `yes' if it does, and `no' if it does not. \\

Example \{i\}: \\
Problem/activity: \\
\{example problem activity\} \\

Concept/skill:  \\
\{example standard description\} \\

Answer:  \\
\{example answer\} \{example thought\} \\

Now, does the problem or activity shown below enable students to learn the full intent of the following concept or skill? \\

Problem/activity:  \\
\{problem activity\} \\

Concept/skill:  \\
\{standard description\} \\

    \end{tcolorbox}
    \caption{Prompt 15 for verification.   
    \label{fig:entailment_15}}
\end{figure*}

\begin{figure*}[h]
    \centering %
    \begin{tcolorbox}[
    taggingPrompt,
    title={\textbf{Prompt 1 for tagging task}},
    ]
    \small
You are a math expert reviewing K-12 curricular materials. You will be shown a problem or activity obtained from school curriculum and a list of one or more \{level\}. Your task is to assign the problem or activity to one or more relevant \{level\} it \{relation\}, and format your output as a comma-separated list of options e.g. ``A, B, C''. \{relation definition\} Output ``none'' if none of the \{level\} below are relevant. DO NOT make up additional \{level\}.\\

Example \{i\}:\\
Problem/activity:\\
\{example problem activity\}\\

\{Level\} options:\\ 
\{example options\}\\

Your response:\\
\{example response\}\\

Now, assign the following problem or activity to one or more relevant \{level\} it \{relation\}.\\

Problem/activity:\\
\{problem activity\}\\

\{Level\} options:\\
\{options\}\\

    \end{tcolorbox}
    \caption{Prompt 1 for tagging.   
    \label{fig:tagging_1}}
\end{figure*}

\begin{figure*}[h]
    \centering %
    \begin{tcolorbox}[
    taggingPrompt,
    title={\textbf{Prompt 3 for tagging task}},
    ]
  \small  
You are a math expert reviewing K-12 curricular materials. You will be shown a problem or activity obtained from school curriculum and a list of one or more \{level\}. Your task is to assign the problem or activity to one or more relevant \{level\} it \{relation\}. \{relation definition\} \\

You should first write a paragraph explaining which \{level\} the problem/activity \{relation\}, and then output a comma-separated list of options. Respond ``none'' if the problem/activity \{relation\} none of the provided \{level\}, and do not make up additional \{level\}. Please format your response in two lines, as shown in the example below:\\

Thought: <your paragraph goes here>\\
Answer: A, C, E\\

Example \{i\}:\\
Problem/activity:\\
\{example problem activity\}\\

\{Level\} options: \\
\{example options\}\\

Thought: \{example thought\}\\
Answer: \{example response\}\\

Now, assign the following problem or activity to one or more relevant \{level\} it \{relation\}.\\

Here is the problem/activity: \\
\{problem activity\}\\

\{Level\} options:\\
\{options\}\\

Please output both your thoughts about what \{level\} this problem \{relation\}, as well as a comma-separated list of \{level\}: \\

    \end{tcolorbox}
    \caption{Prompt 3 for tagging.   
    \label{fig:tagging_3}}
\end{figure*}

\begin{figure*}[h]
    \centering %
    \begin{tcolorbox}[
    taggingPrompt,
    title={\textbf{Prompt 4 for tagging task}},
    ]
   \small 
You are a math instructor reviewing K-12 curricular materials. You will be shown a problem or activity obtained from school curriculum and a list of one or more \{level\}. Your task is to assign the problem or activity to one or more relevant \{level\} it \{relation\}. \{relation definition\}\\

Your response should first begin with a paragraph explaining which \{level\} the problem \{relation\}, and then output a comma-separated list of options. Respond "none" if the problem/activity \{relation\} none of the provided \{level\}. Do not make up additional \{level\}. Please format your response in two lines, as shown in the example below:\\

Thought: <your paragraph goes here>\\
Answer: A, C, E\\

Example \{i\}:\\
Problem/activity:\\
\{example problem activity\}\\

\{Level\} options:\\ 
\{example options\}\\

Your response:\\
Thought: \{example thought\}\\
Answer: \{example response\}\\

Now, assign the following problem or activity to one or more relevant \{level\} it \{relation\}.\\

Problem/activity:\\
\{problem activity\}\\

\{Level\} options:\\
\{options\}\\

Your response:\\

    \end{tcolorbox}
    \caption{Prompt 4 for tagging.   
    \label{fig:tagging_4}}
\end{figure*}

\begin{figure*}[h]
    \centering %
    \begin{tcolorbox}[
    taggingPrompt,
    title={\textbf{Prompt 5 for tagging task}},
    ]
\small
You are a math expert reviewing K-12 curricular materials. You will be shown a problem or activity obtained from school curriculum and a list of one or more \{level\}. Your task is to assign the problem or activity to one or more relevant  \{level\} it  \{relation\}.  \{relation definition\}\\

Your response should be a `json' object with two fields:\\
 \{\\
  ``explanation'': your justification for your answer,
  ``answer'': a succinct comma-separated list of option letters, e.g. ``A, B, C'' \\
 \}\\

If the problem/activity  \{relation\} none of the provided  \{level\}, your answer should be ``none''. Do not make up additional  \{level\}.\\

Example  \{i\}:\\
Problem/activity:\\
 \{example problem activity\}\\

 \{Level\} options: \\
 \{example options\}\\

Your response:\\
 \{\\
  "explanation": "\{example thought\}",\\
  "answer": "\{example response\}"\\
 \}\\

Now, assign the following problem or activity to one or more relevant  \{level\} it  \{relation\}.\\

Problem/activity:\\
\{problem activity\}\\

\{Level\} options:\\
\{options\}\\

Your response:\\
    \end{tcolorbox}
    \caption{Prompt 5 for tagging. Note that the brackets for json formatting serve a different function than the brackets indicating slots in which we input problems/activities, options, and other level-specific information.
    \label{fig:tagging_5}}
\end{figure*}

\begin{figure*}[h]
    \centering %
    \begin{tcolorbox}[
    taggingPrompt,
    title={\textbf{Prompt 6 for tagging task}},
    ]
\small
You are a math expert reviewing K-12 curricular materials. You will be shown a problem or activity obtained from school curriculum and a list of one or more \{level\}. Your task is to assign the problem or activity to one or more relevant \{level\} it \{relation\}. \{relation definition\}\\

Your response should be a `json' object with two fields:\\
\{\\
  "explanation": your reasoning for your answer,\\
  "answer": a succinct comma-separated list of option letters e.g. "A, B, C" \\
\}\\

For example, if the problem or activity \{relation\} both options D and E, the "answer" key would map to "D, E". \\
As another example, if the problem or activity only \{relation\} option A, the "answer" key would map to "A". \\
If the problem/activity \{relation\} none of the provided \{level\}, the "answer" key would map to "none".\\

Do not make up additional \{level\}.\\

Example \{i\}:\\
Problem/activity:\\
\{example problem activity\}\\

\{Level\} options: \\
\{example options\}\\

Your response:\\
\{\\
  "explanation": "\{example thought\}",\\
  "answer": "\{example response\}"\\
\}\\

Now, assign the following problem or activity to one or more relevant \{level\} it \{relation\}.\\

Problem/activity:\\
\{problem activity\}\\

\{Level\} options:\\
\{options\}\\

Your response:

    \end{tcolorbox}
    \caption{Prompt 6 for tagging. Note that the brackets for json formatting serve a different function than the brackets indicating slots in which we input problems/activities, options, and other level-specific information. 
    \label{fig:tagging_6}}
\end{figure*}

\begin{figure*}[h]
    \centering %
    \begin{tcolorbox}[
    taggingPrompt,
    title={\textbf{Prompt 10 for tagging task}},
    ]
\small
You are reviewing K-12 curricular materials. You will be shown a problem or activity obtained from school curriculum and a list of one or more \{level\}. Your task is to assign the problem or activity to one or more relevant \{level\} it \{relation\}. \{relation definition\}\\

Begin your response with a paragraph explaining which \{level\} the problem \{relation\}. You may walk through the act of solving or doing the problem/activity, if possible, to illustrate how it \{relation\} specific \{level\}. Then, conclude with a comma-separated list of options as your answer. Respond "none" if the problem/activity \{relation\} none of the provided \{level\}, and do not make up additional \{level\}. Please format your response in two lines, as shown in the example below:\\

Thought: <your paragraph goes here>\\
Answer: A, C, E\\

Example \{i\}:\\
Problem/activity:\\
\{example problem activity\}\\

\{Level\} options: \\
\{example options\}\\

Your response:\\
Thought: \{example thought\}\\
Answer: \{example response\}\\

Now, assign the following problem or activity to one or more relevant \{level\} it \{relation\}.\\

Problem/activity:\\
\{problem activity\}\\

\{Level\} options:\\
\{options\}\\

Your response:

    \end{tcolorbox}
    \caption{Prompt 10 for tagging.   
    \label{fig:tagging_10}}
\end{figure*}

\begin{figure*}[h]
    \centering %
    \begin{tcolorbox}[
    taggingPrompt,
    title={\textbf{Prompt 11 for tagging task}},
    ]
\small
You are a math expert reviewing K-12 curricular materials. You will be shown a problem or activity obtained from school curriculum and a list of one or more \{level\}. Your task is to assign the problem or activity to one or more relevant \{level\} it \{relation\}. \{relation definition\} \\

You should first write a paragraph explaining which \{level\} the problem/activity \{relation\}, and then output a comma-separated list of options. Respond "none" if the problem/activity \{relation\} none of the provided \{level\}, and do not make up additional \{level\}. Please format your response in two lines. You may walk through how a student may solve or do the problem/activity, if possible, to illustrate how it \{relation\} one or more \{level\}.\\

For example, if the problem or activity \{relation\} \{level\} D and E, your response would be:\\
Thought: <your paragraph goes here>\\
Answer: D, E\\

As another example, if the problem or activity only \{relation\} \{level\} A, your response would be:\\
Thought: <your paragraph goes here>\\
Answer: A\\

Example \{i\}:\\
Problem/activity:\\
\{example problem activity\}\\

\{Level\} options: \\
\{example options\}\\

Your response:\\
Thought: \{example thought\}
Answer: \{example response\}

Now, assign the following problem or activity to one or more relevant \{level\} it \{relation\}.\\

Problem/activity:\\
\{problem activity\}\\

\{Level\} options:\\
\{options\}\\

Your response:

    \end{tcolorbox}
    \caption{Prompt 11 for tagging.   
    \label{fig:tagging_11}}
\end{figure*}

\begin{figure*}[h]
    \centering %
    \begin{tcolorbox}[
    taggingPrompt,
    title={\textbf{Prompt 12 for tagging task}},
    ]
\small
You are a math instructor reviewing K-12 curricular materials. You will be shown a problem or activity obtained from school curriculum and a list of one or more \{level\}. Your task is to assign the problem or activity to one or more relevant \{level\} it \{relation\}. \{relation definition\}\\

Your response should be a `json' object with two fields: "explanation", which includes your reasoning, and "answer", which is a comma-separated list of option letters. For example: \\
\{\\
  "explanation": <your reasoning goes here>,\\
  "answer": "A, B, C"\\
\}\\

If the problem/activity \{relation\} none of the provided \{level\}, your answer should be "none". To help you justify your answer, you may try solving the problem or doing the activity, if possible.\\

Do not make up additional \{level\}.\\

Example \{i\}:\\
Problem/activity:\\
\{example problem activity\}\\

\{Level\} options: \\
\{example options\}\\

Your response, in json format:\\
\{\\
  "explanation": "\{example thought\}",\\
  "answer": "\{example response\}"\\
\}\\

Now, assign the following problem or activity to one or more relevant \{level\} it \{relation\}.\\

Problem/activity:\\
\{problem activity\}\\

\{Level\} options:\\
\{options\}\\

Your response, in json format:\\

    \end{tcolorbox}
    \caption{Prompt 12 for tagging. Note that the brackets for json formatting serve a different function than the brackets indicating slots in which we input problems/activities, options, and other level-specific information.}
    \label{fig:tagging_12}
\end{figure*}

\end{document}